\documentclass[sigconf]{aamas} 

\usepackage{balance} 
\usepackage{wrapfig}

\setcopyright{ifaamas}
\acmConference[AAMAS '24]{Proc.\@ of the 23rd International Conference
on Autonomous Agents and Multiagent Systems (AAMAS 2024)}{May 6 -- 10, 2024}
{Auckland, New Zealand}{N.~Alechina, V.~Dignum, M.~Dastani, J.S.~Sichman (eds.)}
\copyrightyear{2024}
\acmYear{2024}
\acmDOI{}
\acmPrice{}
\acmISBN{}

\acmSubmissionID{755}

\title[AAMAS-2024 Formatting Instructions]{Surge Routing: Event-informed Multiagent Reinforcement Learning for Autonomous Rideshare}

\author{Daniel Garces}
\affiliation{
  \institution{Harvard University}
  \city{Boston, MA}
  \country{United States}}
\email{dgarces@g.harvard.edu}

\author{Stephanie Gil}
\affiliation{
  \institution{Harvard University}
  \city{Boston, MA}
  \country{United States}}
\email{sgil@seas.harvard.edu}

\begin{abstract}
  Large events such as conferences, concerts and sports games, often cause surges in demand for ride services that are not captured in average demand patterns, posing unique challenges for routing algorithms. We propose a learning framework for an autonomous fleet of taxis that leverages event data from the internet to predict demand surges and generate cooperative routing policies. We achieve this through a combination of two major components: (i) a demand prediction framework that uses textual event information in the form of events' descriptions and reviews to predict event-driven demand surges over street intersections, and (ii) a scalable multiagent reinforcement learning framework that leverages demand predictions and uses one-agent-at-a-time rollout combined with limited sampling certainty equivalence to learn intersection-level routing policies.
  For our experimental results we consider real NYC ride share data for the year 2022 and information for more than 2000 events across 300 unique venues in Manhattan. We test our approach with a fleet of 100 taxis on a map with 2235 street intersections. Our experimental results demonstrate that our method learns routing policies that reduce wait time overhead per serviced request by $25\%$ to $75\%$, while picking up $1\%$ to $4\%$ more requests than other model-based RL frameworks and classical methods in operations research.
\end{abstract}


\newcommand{\BibTeX}{\rm B\kern-.05em{\sc i\kern-.025em b}\kern-.08em\TeX}

\begin{document}


\pagestyle{fancy}
\fancyhead{}


\maketitle 


\section{Introduction}
\begin{wrapfigure}[13]{r}{0.44\linewidth}
    \centering
    \includegraphics[width=\linewidth]{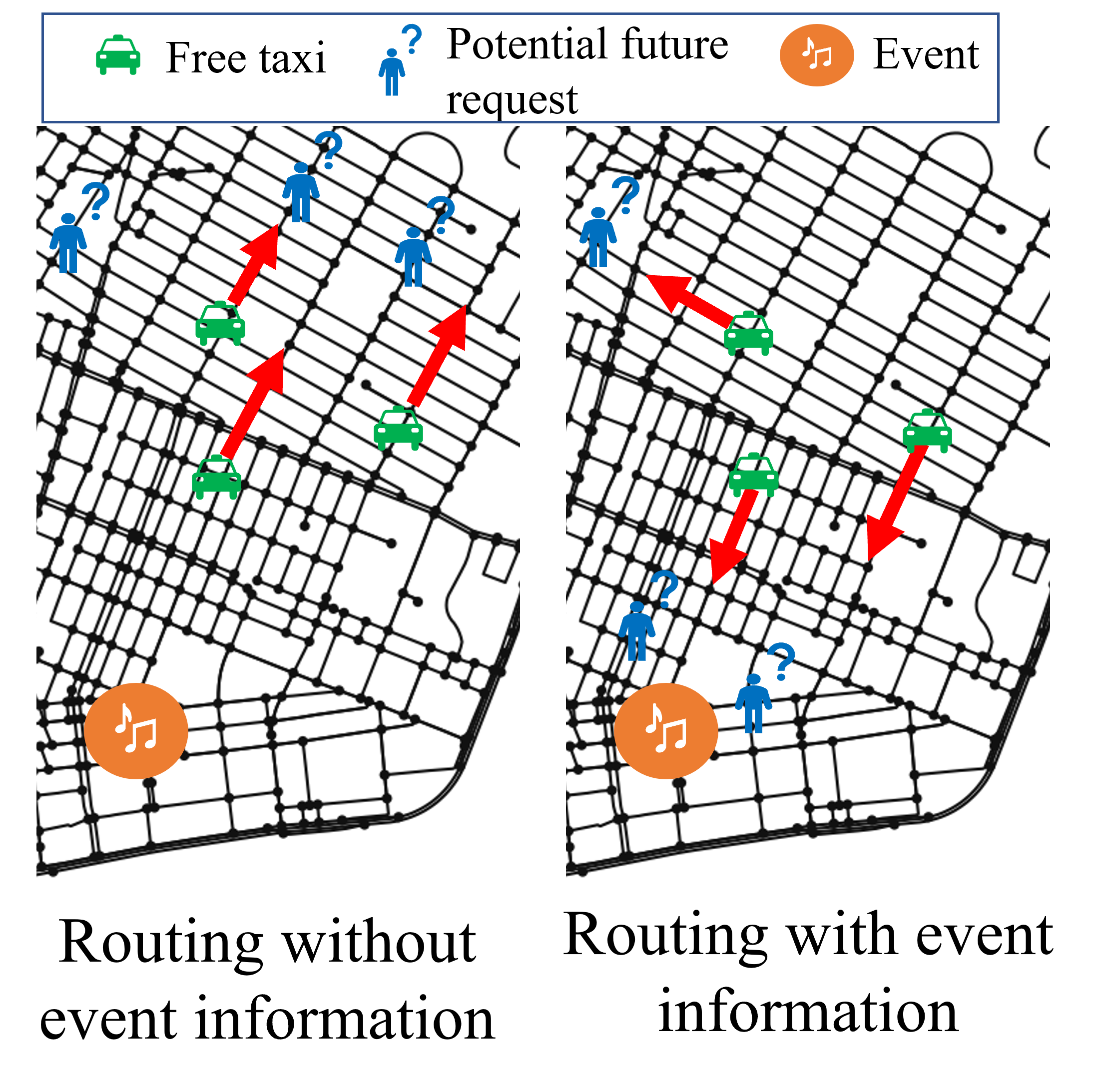}
    \vspace{-20pt}
    \caption{\small{Motivating example}}
    \label{fig:motivation}
    \Description{Motivating example showing that taxis that do not consider event-driven demand surges get routed away from events based on their inaccurate demand estimation, while routing protocols that do consider event-driven demand surges are able to route taxis towards event venues preemptively.}
\end{wrapfigure}

Large events such as conferences, concerts, and sports games lead to a large agglomeration of people and hence tend to produce demand surges for ride services. Efficiently servicing these surges requires the orchestration of fleet-wise coordinated plans that leverage accurate estimates of event-driven demand fluctuations (see Fig~\ref{fig:motivation}). Unfortunately, most on-demand mobility routing algorithms in the literature do not address event-driven demand surges, either because they do not consider potential future requests and thus do not plan ahead \citep{Karp1990, Duan2014, Bertsimas2019OnlineVR, Bertsekas1979Auction, Croes1958, LSCP2003, Mitchell1998}, or because they plan ahead using demand models that are based on averaged or short-term history and hence do not account for point disturbances associated with event-related surges~\citep{Ulmer2018, Farhan2018, Chen2020, Chng2022, AHAMED2021227, PARVEZFARAZI2021100425, Delarue2020, Nazari2018, garces2023}.

The plethora of event data freely available on the internet and recent advances in language models allow for the development of event-informed demand prediction mechanisms \citep{Markou2019, Markou2020, Rodrigues2019}. Additionally, recent advances in multiagent Reinforcement Learning (RL) \citep{gammelli2022graph, enders2023hybrid}, including rollout-based mechanisms \citep{bertsekas2020rollout, Bertsekas2022AlphaZero, garces2023}, allow for learning multiagent cooperative plans. Ideally, demand prediction mechanisms and RL-based routing algorithms could be combined to obtain a fine-grained event-informed multiagent routing framework that predicts event-driven demand surges and routes taxis accordingly. However, designing and implementing such a framework for a city-scale application is a non-trivial task. Tackling this task requires addressing two major challenges:
(i) demand predictions should leverage data from multiple events to generate accurate estimates of future demand, that additionally, are usable by intersection-level multiagent RL routing algorithms and (ii) the expected cost of agents' actions should be approximated by the multiagent RL routing algorithm at a city-scale without incurring prohibitively long execution times.

In this paper we address these two challenges by introducing a multiagent routing framework that leverages event-informed hourly demand predictions to learn cooperative routing policies on a city-scale environment. To our knowledge, our proposed method is the first work to integrate event-driven surge demand prediction and intersection-level multiagent routing. 
Our approach solves the first challenge by leveraging an unsupervised aggregation mechanism that we developed. This mechanism aggregates data from several overlapping events over city blocks (or ``sectors'') to generate sector-level demand predictions. These demand predictions are then mapped to intersections using a probabilistic assignment mechanism that we design. In this way, demand predictions over intersections can then be used to inform future cost estimation for our RL routing algorithm. Since the state space for a city-scale application is very large, standard rollout methods tend to need a large number of samples to approximate the expected future costs of actions. For this reason,  our approach solves the second challenge by leveraging certainty equivalence approximation, where the demand distribution is replaced by a semi-deterministic mean value. We also use a limited sampling modification to the standard certainty equivalence to reduce the sampling space to the surrounding area of a given sector when estimating future costs for an agent inside that sector. This modification allows our method to reduce the number of samples required for estimating the expected costs of actions while still maintaining similar accuracy. We are able to maintain accuracy since taxis that are very far away from a sector can't reach the sector during their planning horizon and hence have very little effect on that sector's planning.

\begin{figure*}[t]
    \centering
    \includegraphics[width=0.7\textwidth]{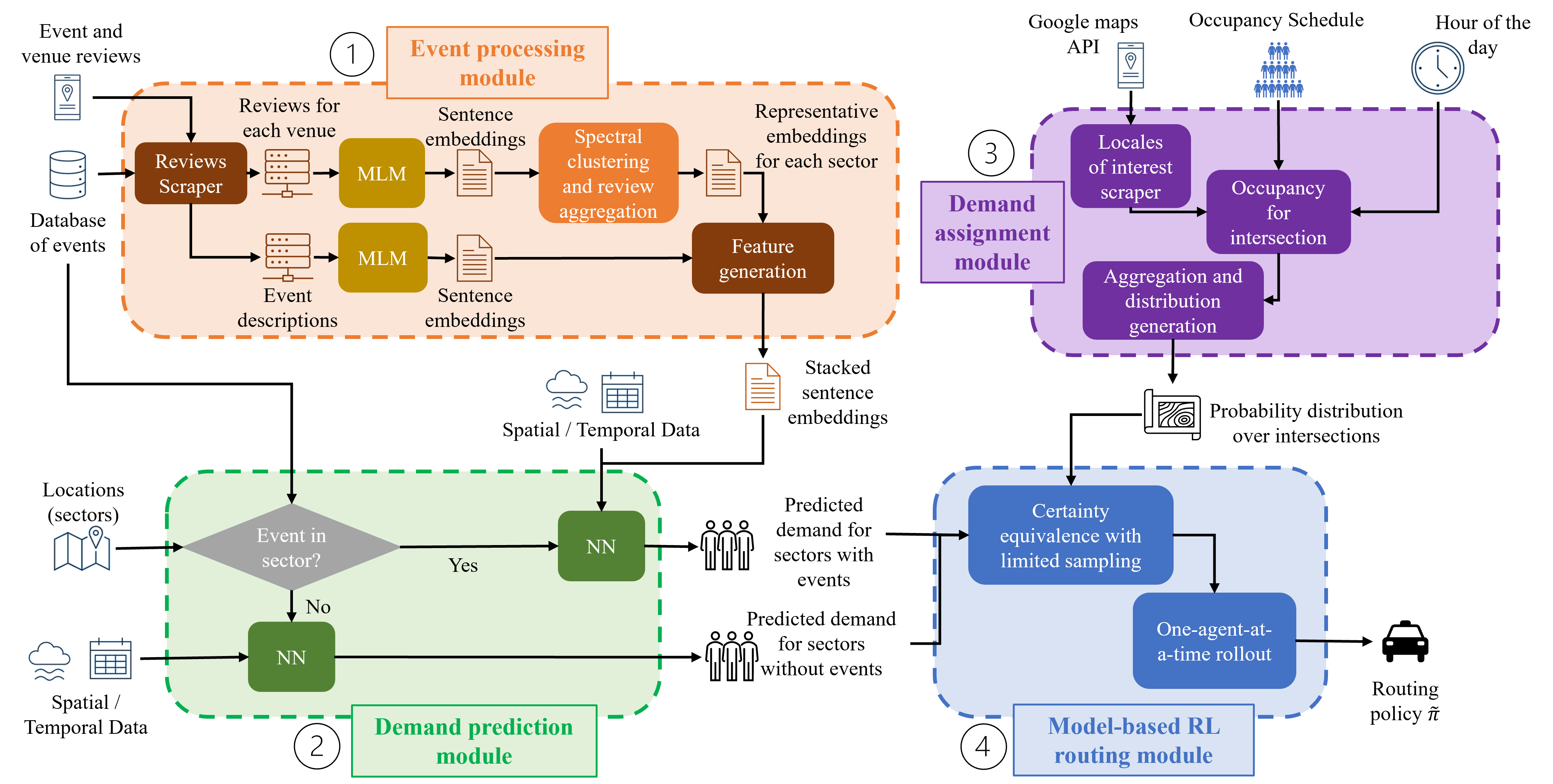}
    \caption{\small{General System Overview showing our proposed approach's four modules: 1) the event processing module, 2) the demand prediction module, 3) the demand assignment module, and 4) the model-based RL routing module.}}
    \label{fig:general_overview}
    \Description{Graphical representation of the overall structured of our proposed approach. Our approach is composed of four modules: 1) the event processing module, 2) the demand prediction module, 3) the demand assignment module, and 4) the model-based RL routing module}
\end{figure*}

More specifically, our proposed framework is composed of four modules as shown in Fig.~\ref{fig:general_overview}: 1) the event processing module, 2) the demand prediction module, 3) the demand assignment module, and 4) the model-based RL routing module. The event processing module captures event information from the internet in the form of event reviews and descriptions by leveraging sentence embeddings \citep{conneau2018supervised, subramanian2018learning, cer2018universal, barkan2019scalable, yang-etal-2021-universal, Liu2019RoBERTaAR} generated from a pre-trained Masked Language Model (MLM). Sentence embeddings for events in overlapping sectors of the map are aggregated using an unsupervised aggregation scheme. This aggregation scheme combines spectral clustering \citep{Karfi2018, zhu2014tripartite, Damle2018} and graph summarization \citep{erkan2004university, zhu2013graph} to produce representative dense vectors for each sector. These dense vectors are then used by the demand prediction module to predict hourly demand for the taxi service at each sector. In order to integrate the demand prediction into the multiagent RL-based routing mechanism, we propose a novel demand assignment module. This module uses a probabilistic assignment routine that leverages locale maximum occupancy data and occupancy schedules (percentage of maximum occupancy)\citep{Happle2019, PANCHABIKESAN2021119539, ASHRAE2013, Comnet2016} to map demand predictions over sectors to demand predictions over street intersections. These demand predictions over intersections are then used by the multiagent routing scheme to estimate expected future costs for each agent's actions. Our proposed multiagent routing scheme builds off one-agent-at-a-time rollout \citep{BERTSEKAS2020Multiagent, Bertsekas2021PI, Bertsekas2022AlphaZero, garces2023} and combines it with a limited certainty equivalence approximation to reduce the sampling complexity of estimating expected future costs. 

We test our approach using real NYC's Taxi and Limousine High-Volume For-Hire-Vehicle (HV-FHV) data \cite{TLC_Data} and information for more than $2000$ events across $300$ venues. We consider a region of Manhattan with $2235$ intersections and a fleet of $100$ autonomous taxis. This setup has 3X more intersections and 6X more taxis than most existing works in this area, demonstrating the scalability of our approach. We empirically demonstrate that our event-informed framework learns routing policies that reduce wait time overhead per serviced request by $25\%$ to $75\%$ while picking up $1\%$ to $4\%$ more requests than other model-based RL frameworks and other classical algorithms in operations research.
\footnote{We define wait time overhead as the additional time that a request will have to wait in a realistic stochastic setting. In contrast to a setting in which the locations and entry times of all requests are known a-priori, taxis cannot be routed to the exact locations of requests from the beginning of the episode.} 

\section{Related Works}
In this section, we review the current state-of-the-art for the two problems related to event-informed multiagent routing: Dynamic Vehicle Routing (DVR), and demand prediction.

\paragraph{Dynamic Vehicle Routing} 
Earlier works tackled this problem using instantaneous assignment approaches \cite{Karp1990, Duan2014, Bertsimas2019OnlineVR, Bertsekas1979Auction}, and routing heuristics, \cite{Croes1958, LSCP2003, Mitchell1998}. Instantaneous assignment approaches, however, produce myopic policies since they do not consider potential future requests. Sampling-based stochastic optimization \cite{LOWALEKAR201871} try to solve this issue, but incur long computation times due to the multistep planning objective and large state space. To improve computation times, several authors considered offline trained approximations \cite{Ulmer2018, AHAMED2021227, PARVEZFARAZI2021100425, Delarue2020, Nazari2018}.  Offline learning methods tend to be computationally faster at inference time, but they do not generalize to unknown scenarios not represented in the training data, and tend to not scale well as the state space becomes larger. This makes them infeasible for deployment in city-scale urban environments, where the state space is very large and new changes in demand are not necessarily represented in the historical demand data. To try to address this issue, and allow policies to adapt to changes in the demand, other authors have considered online optimization methods \cite{SiV10, Bent2004a, bertsekas2020rollout, Bertsekas2021PI}, and hybrid approaches \cite{garces2023}. These online learning methods, however, do not consider event-driven surges in demand, and hence they tend to not be applicable to realistic urban settings. In this paper we aim to address this limitation by integrating event-informed demand prediction into model-based RL routing.

\paragraph{Demand prediction} 
Some authors have considered time series analysis techniques \cite{miao2017datadriven, Moreira2013, Li2012}. These approaches, however, rely on the data being stationary or following predictable seasonal changes that can easily be removed by seasonal-differencing. Demand for transportation services, however, tends to be highly dynamic and it is affected by external events like concerts or traffic accidents. This situation prevents time series analysis methods from performing as expected. To address this shortcoming, various authors have considered learning approaches \cite{Liu2019, Chen2021UberNet, Nejadettehad2019, Cao2022} to predict demand based on spatial and temporal features (time of day, weather conditions, etc.). However, these methods generally do not employ information about events, and hence do not generalize to scenarios where events cause demand surges. Other authors also look at predicting demand using spatial and temporal features, as well as event information \cite{Markou2019, Rodrigues2019, Markou2020}. One limitation of these methods that we address in this paper is the ability to include information from an arbitrary number of events over overlapping regions of the map. The ability of our method to aggregate event information makes it applicable to city-scale environments where there are multiple venues hosting simultaneous events in the same region of the map.


\section{Problem Formulation}
In this section, we present the formulation for our multiagent taxicab routing problem, casting it as a city-scale discrete time, finite horizon, stochastic Dynamic Programming (DP) problem. We define the environment, requests, state and control space, the basics of rollout and our problem of interest in the following subsections.
\label{sec:problem_formulation}

\subsection{Environment}
\begin{wrapfigure}[15]{l}{0.25\linewidth}
    \centering
    \includegraphics[width=\linewidth]{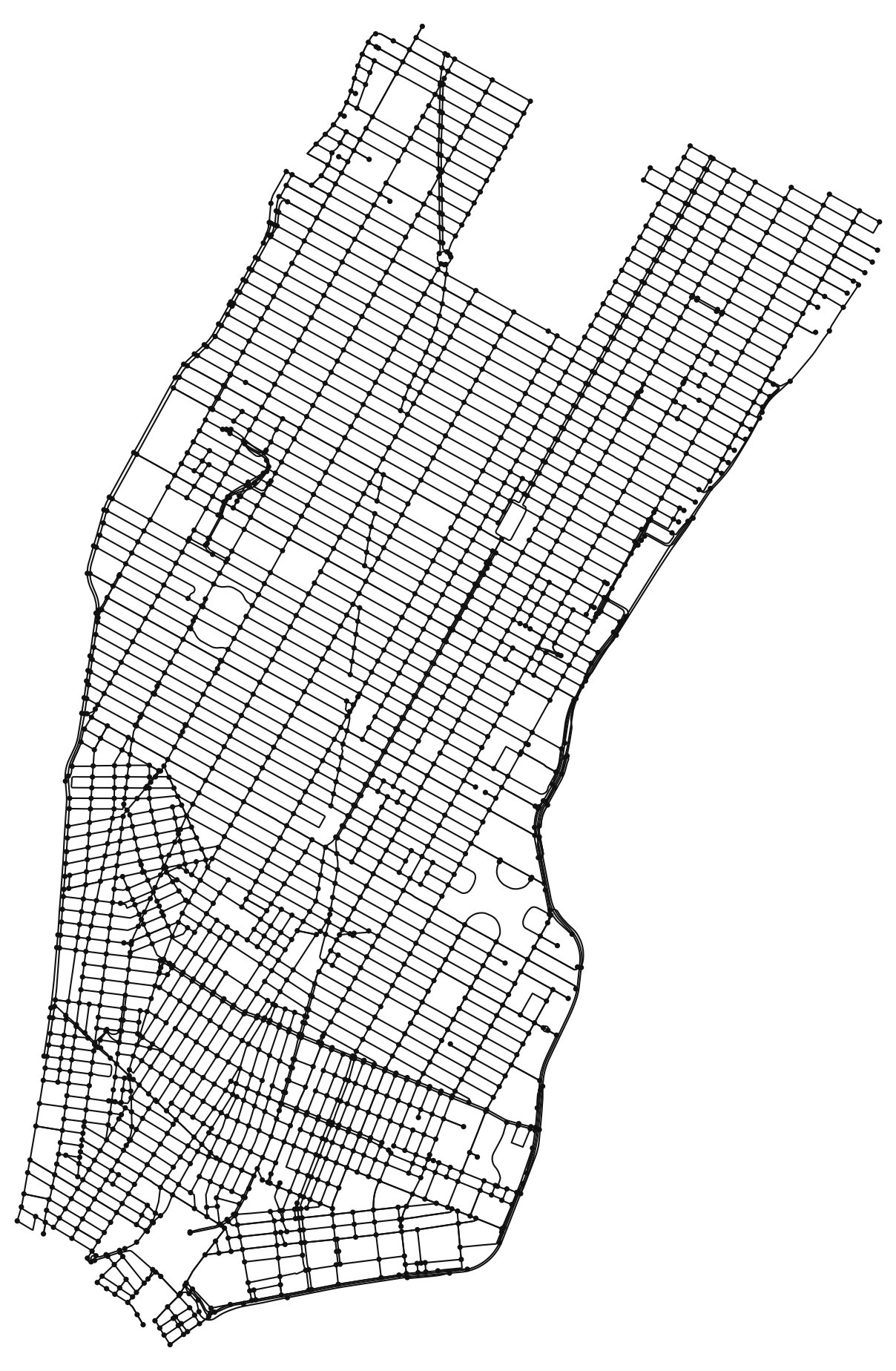}
    \vspace{-20pt}
    \caption{\small{Our routing environment over 2235 intersections in NYC}}
    \label{fig:example_env}
    \Description{Graphical representation of the NYC street network that corresponds to our routing environment. This street network has 2235 intersections.}
\end{wrapfigure}

We assume that taxicabs are deployed in an urban environment with a fixed street topology (see Fig.~\ref{fig:example_env}). The environment is expressed as a graph $G=(V, E)$, where $V=\{ 1, \dots, n\}$ corresponds to the set of intersections in the map numbered $1$ through $n$, while $E \subseteq \{ (i,j) | i, j \in V\}$ corresponds to the set of directed streets that connect intersections $i$ and $j$ in the map. The set of adjacent intersections to intersection $i$ is denoted as $\mathcal{N}_i = \{ j | j\in V, (i,j) \in E\}$. We also assume that the environment is divided into sectors $s_k \subseteq V$, such that $V = \bigcup_k s_k$ and $s_k \cap s_h = \emptyset, \forall k \neq h$. We denote the set of all sectors in the map as the set $S$, where $S$ is then a set of sets. We define $\mathcal{A}(s_k) \subseteq S$ as the set of all the sectors adjacent to sector $s_k$. We define $\mathcal{I}_{s_k} \subseteq s_k$ as the set of intersections in sector $s_k$ that can be used as pickup or drop-off locations, following local regulations. We denote the set of intersections that can be used for pickup or drop-off of requests over the entire map as $\mathcal{I}_{V} = \bigcup_{s_k \in S} \mathcal{I}_{s_k}$.

\subsection{Requests}
We define a request $r$ for the ride service as a tuple $r = \left< \rho_r, \delta_r, t_r, \phi_r \right>$, where $\delta_r \in \mathcal{I}_{V}$ and $\rho_r \in \mathcal{I}_{V}$ correspond to the nearest available intersection to the request's desired pickup and drop-off locations, respectively; $t_r$ corresponds to the time at which the request entered the system; and $\phi_r \in \{ 0, 1\}$ is an indicator, such that $\phi_r = 1$ if the request has been picked up by a vehicle, $\phi_r = 0$ otherwise. We model the number of new pickup requests for a specific sector $s_k$ as a random variable $\eta_{s_k}$ with an unknown underlying distribution $p_{\eta_{s_k}}$. Its estimation is denoted as $\tilde{p}_{\eta_{s_k}}$. We denote the realization of $\eta_{s_k}$ at time $t$ as $\eta_{s_k}(t)$. We denote the set of new pickup requests at time $t$ for a specific sector $s_k$ as $\mathbf{r_{s_k, t}} = \{ r | \rho_r \in  \mathcal{I}_{s_k}, t_r = t\}$, such that $|\mathbf{r_{s_k, t}}| = \eta_{s_k}(t)$. We model the exact pickup intersections for an arbitrary request given that the request originated at sector $s_k$ as the random variable $\rho_{s_k}$ with support $\mathcal{I}_{s_k}$. Similarly, we model the exact drop-off intersection for an arbitrary request given that the request will be dropped off at sector $s_k$ as the random variable $\delta_{s_k}$ with support $\mathcal{I}_{s_k}$. Both of these random variables have unknown underlying distributions that we denote as $p_{\rho_{s_k}}$ and $p_{\delta_{s_k}}$, respectively. We also denote their estimated probability distributions as $\tilde{p}_{\rho_{s_k}}$ and $\tilde{p}_{\delta_{s_k}}$, respectively. We model the drop-off sector for an arbitrary request given that the request has pick up sector $s_k$ as the random variable $\beta_{s_k}$ with an unknown probability distribution $p_{\beta_{s_k}}$. We denote its estimated probability distribution as $\tilde{p}_{\beta_{s_k}}$. We denote the demand model for a given sector $s_k$ as the set of random variables $D_{s_k} = \{ \eta_{s_k}, \rho_{s_k}, \delta_{s_k}, \beta_{s_k} \}$. We denote the global demand model as $\mathcal{D} = \bigcup_{s_k \in S} D_{s_k}$. We define $\mathbf{\overline{r}_{s_k, t}} = \{ r | r \in \mathbf{r_{s_k, t'}}, \phi_r=0, t' = 1, \dots t \}$ as the set of outstanding pickup requests for sector $s_k$ that have not been picked up by any taxi till time $t$. We denote the set of all outstanding pickup requests at time $t$ as $\mathbf{\overline{r}_{t}} = \bigcup_{s_k \in S} \mathbf{\overline{r}_{(s_k, t)}}$

\subsection{State representation and control space}
We assume there is a total of $m$ agents and all agents can perfectly observe all available requests, and all other agents' locations and occupancy status. We assume that this $m$ is fixed over time. We represent the state of the system at time $t$ as a tuple $x_t = \left< \vec{\nu_t}, \vec{\tau_t}, \mathbf{\overline{r}_{t}} \right>$. $\vec{\nu_t}=[\nu_t^1, \dots, \nu_t^m]$ corresponds to the list of locations for all $m$ agents at time $t$, where $\nu_t^\ell \in V$ corresponds to the closest intersection to the geographical position of agent $\ell$. $\vec{\tau_t} = [\tau_t^1, \dots, \tau_t^m]$ is the list of time remaining in current trip for all $m$ agents. If agent $\ell$ is available, then it has not picked up a request and hence  $\tau_t^\ell = 0$, otherwise $\tau_t^\ell \in \mathbb{N}^+$.

We denote the control space for agent $\ell$ at time $t$ as $\mathbf{U}_t^\ell(x_t)$. If the agent is available (i.e. $\tau_t^\ell = 0$), then $\mathbf{U}_t^\ell(x_t) = \{ \mathcal{N}_{\nu_t^\ell}, \nu_t^\ell, \psi\}$, where $\mathcal{N}_{\nu_t^\ell}$ corresponds to the set of adjacent intersections to the current location $\nu_t^\ell$, and $\psi$ corresponds to a special pickup control that becomes available if there is a request $r \in \mathbf{\overline{r}_{t}}$ a the location of agent $\ell$ such that $\rho_r = \nu_t^\ell$. If the agent is currently servicing a request $r$ (i.e. $\tau_t^\ell > 0$), then $\mathbf{U}_t^\ell(x_t) = \{ \zeta \}$, where $\zeta$ corresponds to the next hop in Dijkstra's shortest path between agent $\ell$'s current location $\nu_t^\ell$ and the destination of the request $\delta_r$. We denote all possible controls for all $m$ agents at state $x_t$ as $\mathbf{U}_t (x_t) = \mathbf{U}_t^1(x_t) \times \dots \times \mathbf{U}_t^m(x_t)$. 

\subsection{Rollout-based routing}
\label{subsec:rollout}
We are interested in learning a cooperative pickup and routing policy that minimizes the total wait for all passengers over a finite time horizon of length $N$. We denote the state transition function as $f$, such that $x_{t+1} = f_t\left(x_t, u_t, \mathcal{D} \right)$, where $x_{t+1}$ is the resulting state after control $u_t \in \mathbf{U}_t (x_t)$ has been applied from state $x_t$ considering the realizations for all random variables in $\mathcal{D}$. We define the stage cost $g_t\left(x_t, u_t, \mathcal{D} \right) = |\mathbf{\overline{r}_{t}}|$ as the number of outstanding requests over the entire map at time $t$. We define a policy $\pi = \{ \mu_1, \dots \mu_N\}$ as a set of functions that maps state $x_t$ into control $u_t = \mu_t(x_t) \in \mathbf{U}_t (x_t)$, with its cost given by $J_{\pi}(x_1) = E \left[g_N(x_N) + \sum_{t=1}^{N-1}g_t\left(x_t, \mu_t(x_t), \mathcal{D} \right) \right]$, where $g_N(x_N) = |\overline{\mathbf{r_{N}}}|$ is the terminal cost. Since the control space is a cartesian product and hence it grows exponentially with the number of agents, searching the entire control space to find the optimal policy is computationally intractable. For this reason, we consider policy improvement schemes, such as rollout \citep{BERTSEKAS2020Multiagent, Bertsekas2021PI}, that allows us to obtain a lower cost policy by improving upon a base policy with a reasonable initial behavior. We define a base policy $\bar{\pi} = \{ \bar{\mu}_1, \dots \bar{\mu}_N \}$ as an easy to compute heuristic that is given. Our objective becomes then to find an approximate policy $\tilde{\pi} = \{ \tilde{\mu}_1, \dots \tilde{\mu}_N \}$, such that given base policy $\bar{\pi}$ the minimizing action for state $x_t$ at time $t$ is given by:
\begin{equation}
    \tilde{\mu}_t(x_t) \in \arg \min_{u_t \in \mathbf{U}_t (x_t)} E_{\mathcal{D}} \left[ g_t\left(x_t, u_t, \mathcal{D} \right) + \tilde{J}_{\bar{\pi},t}(x_{t+1}) \right]
    \label{eq:Bellman_eq}
\end{equation}
Where $\tilde{J}_{\bar{\pi},t}(x_{t+1}) = \hat{J}_{T} + \sum_{t' = (t+1)}^{T} g_{t'}(x_{t'}, \bar{\mu}_{t'}(x_{t'}), \mathcal{D})$ is a cost approximation derived from applying the base policy $\bar{\pi}$ for $H$ time steps from state $x_{t+1}$ with a terminal cost approximation $\hat{J}_{T} = |\mathbf{\overline{r}_{T}}|$, where $T=(t+1) + H$. The expectation in eq.~\ref{eq:Bellman_eq} is estimated using Monte-Carlo simulations.

\subsection{Problem}
\label{subsec:problem} 
Recent success of rollout methods in combinatorial applications \citep{bertsekas2020rollout, Bhattacharya2020MultiagentRollout, Bertsekas2022AlphaZero}, including small scale routing \citep{garces2023}, makes these rollout-based methods a natural choice for tackling city-scale routing. However, applying rollout methods to a city-scale environment with event-driven demand surges is a non-trivial task. To integrate the rollout formulation presented in Sec.~\ref{subsec:rollout} into our city-scale application, we need to solve two Problems. \\
\textbf{Problem 1}: We must estimate the probability distributions of the random variables that specify the demand model $\mathcal{D}$ to obtain $\tilde{\mathcal{D}} = \bigcup_{s_k \in S} \{ \tilde{\eta}_{s_k}, \tilde{\rho}_{s_k}, \tilde{\delta}_{s_k}, \tilde{\beta}_{s_k} \}$ with underlying distributions $\tilde{p}_{\eta_{s_k}}$, $\tilde{p}_{\rho_{s_k}}$, $\tilde{p}_{\delta_{s_k}}$, and $\tilde{p}_{\beta_{s_k}}$ that capture demand surges produced by events. \\
\textbf{Problem 2:} We must reduce the computational cost of approximating the expectation Eq.~\ref{eq:Bellman_eq}, for the multiagent case, in order to make rollout-based methods amenable to city-scale environments.

\section{Our Approach}
To solve \textbf{Problem 1} in Sec.~\ref{subsec:problem}, we need to derive estimates $\tilde{p}_{\eta_{s_k}}$, $\tilde{p}_{\rho_{s_k}}$, $\tilde{p}_{\delta_{s_k}}$, and $\tilde{p}_{\beta_{s_k}}$. To estimate $\tilde{p}_{\eta_{s_k}}$, we develope a novel \emph{event-informed demand estimation procedure} that leverages sentence embeddings \citep{Liu2019RoBERTaAR} and spectral clustering techniques\citep{Damle2018} to generate vector representations for events. We propose a demand prediction scheme that predicts the number of requests that will enter the system at each sector $s_k \in S$ using these vector representations. Finally, We build off of the idea of Certainty Equivalence \citep{BertsekasCE} to obtain a semi-deterministic approximation for $\tilde{p}_{\eta_{s_k}}$ by uniformly distributing the predicted number of requests for sector $s_k$ over the entire time horizon $N$. To estimate $\tilde{p}_{\rho_{s_k}}$ and $\tilde{p}_{\delta_{s_k}}$, we leverage occupancy schedules \citep{Comnet2016, ASHRAE2013} to derive a novel probabilistic method that maps demand over a sector $s_k$ to individual intersections $j\in \mathcal{I}_{s_k}$. To estimate $\tilde{p}_{\beta_{s_k}}$, we use the relative frequency of pickup and drop-off sectors in historical data conditioned on having pickup sector $s_k$. The combination of all these estimation procedures, allows our method to account for demand surges and transfer that information to our routing framework.

To solve \textbf{Problem 2} in Sec.~\ref{subsec:problem}, we build off one-agent-at-a-time rollout \citep{BERTSEKAS2020Multiagent, Bertsekas2021PI} and Certainty Equivalence \citep{BertsekasCE}. We develop a novel rollout-based scalable routing framework that scales linearly with the number of agents and has a reduced sampling space. This allows our system to decrease the number of simulations required for the Monte-Carlo estimation of the expectation in Eq.~\ref{eq:Bellman_eq}. The combination of one-agent-at-a-time rollout and our novel limited Certainty Equivalence approximation makes our system capable of handling large fleets of taxis over large maps without incurring prohibitively long computation times.

We will cover the specifics of our scalable routing framework first, in order to motivate the need for the estimation of $\tilde{p}_{\eta_{s_k}}$, $\tilde{p}_{\rho_{s_k}}$, $\tilde{p}_{\delta_{s_k}}$, and $\tilde{p}_{\beta_{s_k}}$.

\subsection{Rollout-based scalable routing framework}
\label{subsec:routing_framework}
\begin{figure}
    \centering
    \includegraphics[width=0.72\linewidth]{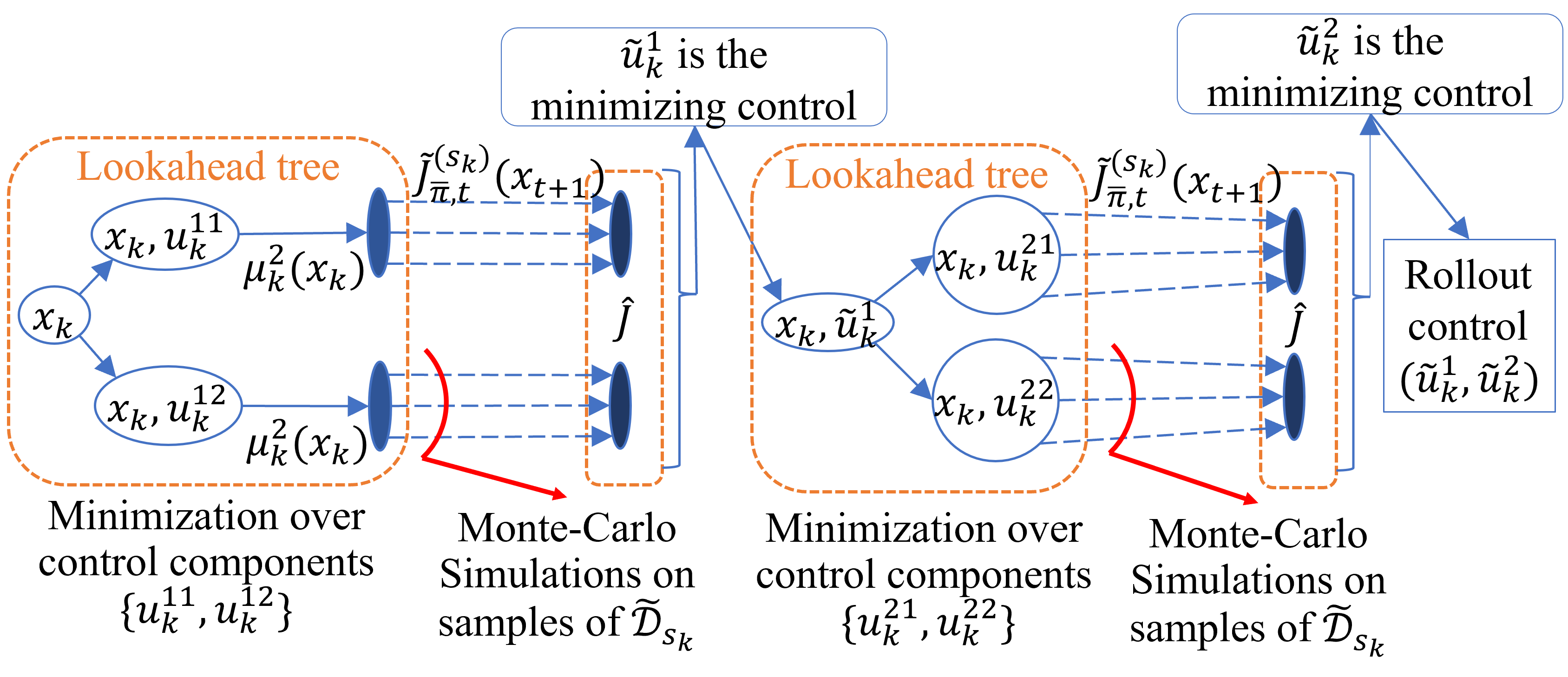}
    \vspace{-5pt}
    \caption{\small{Example of one-agent-at-a-time rollout with two agents, where each agent only has two available actions.}}
    \Description{Pictorial representation of one-agent-at-a-time rollout in an environment with two agents, where each agent only has two available actions. One-agent-at-a-time rollout is composed of a lookahead step that expands all possible actions for one agent and then a truncated rollout that estimates the future cost of executing that action by obtaining Monte-Carlo samples from the demand distributions. The action for each agent is chosen based on its expected cost (each agent chooses the action with the smallest cost).}
    \label{fig:oat_example}
\end{figure}
The proposed rollout-based scalable routing framework is mostly composed of the model-based RL module shown in Fig.~\ref{fig:general_overview}.4, which leverages one-agent-at-a-time rollout \cite{bertsekas2020rollout},\cite{Bhattacharya2020MultiagentRollout}, combined with a limited sampling formulation derived from the idea of scenarios in certainty equivalence \citep{BertsekasCE} to obtain policy $\tilde{\pi}$. We choose one-agent-at-a-time rollout as the foundation for our algorithm since the control space for this rollout variation scales linearly with the number of agents, instead of exponentially \citep{bertsekas2020rollout}. Under our proposed framework, we replace the minimization (Eq.~\ref{eq:Bellman_eq}) given in problem statement for a one-agent-at-a-time formulation. More specifically, agent $\ell$'s one-at-a-time rollout control at state $x_t$, given that agent $\ell$ is at an intersection $j$ that is inside sector $s_k$, is then:
\begin{equation}
    \tilde{u}_t^{\ell} \in   \arg \min_{u^\ell_t \in \mathbf{U}^\ell_k(x_t)} E_{\tilde{\mathcal{D}}_{s_k}}[ g_t(x_t, {\bar{u}}, \tilde{\mathcal{D}}_{s_k}) + \tilde{J}_{\bar{\pi}, t}^{(s_k)}(x_{t+1})]
     \label{eq:OAT-rollout}
\end{equation}
where ${\bar{u}}=(\tilde{u}_t^{1},\ldots, \tilde{u}_t^{\ell-1}, u^\ell_t, \bar{\mu}_t^{\ell+1}(x_t),\ldots, \bar{\mu}_t^{m}(x_t))$; the variable $\tilde{\mathcal{D}}_{s_k} = \{ \tilde{\eta}_{s_h}, \tilde{\rho}_{s_h}, \tilde{\delta}_{s_h}, \tilde{\beta}_{s_h} | \forall s_h \in \mathcal{A}(s_k) \bigcup \{ s_k\} \}$ corresponds to the \emph{local} set of random variables of \emph{adjacent sectors} to $s_k$ where each random variable has estimated probability distributions derived from the \emph{event-informed demand estimation procedure}; and $\tilde{J}_{\bar{\pi}, t}^{(s_k)}(x_{t+1}) = \hat{J}_{T} + \sum_{t' = (t+1)}^{T} g_{t'}(x_{t'}, \bar{\mu}_{t'}(x_{t'}), \tilde{\mathcal{D}}_{s_k})$ corresponds to the cost approximation of executing the base policy for $H$ steps, considering samples from the estimated demand distributions in $\tilde{\mathcal{D}}_{s_k}$, and a terminal cost approximation $\hat{J}_{T} = |\mathbf{\overline{r}_{T}}|$ with $T=(t+1)+H$. The expectation in Eq.~\ref{eq:OAT-rollout} is estimated using Monte-Carlo approximation. A graphical example of the proposed formulation is presented in Fig.~\ref{fig:oat_example}. By considering only the current and adjacent sectors when obtaining samples for the Monte-Carlo approximation of the expected cost of each potential action for a given agent we are able to decrease the sample space and reduce the computational complexity of the sampling procedure for the estimation of the expectation. If we combine this approach with certainty equivalence scenarios \citep{BertsekasCE}, where we consider a finite number of representative scenarios instead of a full stochastic set of samples, the resulting system incurs lower execution times when approximating the expectation in Eq.~\ref{eq:OAT-rollout}, while still obtaining an accurate approximation.

\subsection{Event-informed demand estimation}
\label{subsec:event-informed_demand_estimation}
In this section we explain how we estimate the underlying probability distributions $\tilde{p}_{\eta_{s_k}}$, $\tilde{p}_{\rho_{s_k}}$, $\tilde{p}_{\delta_{s_k}}$, and $\tilde{p}_{\beta_{s_k}}$, which compose $\tilde{\mathcal{D}}_{s_k}$ the demand model for sector $s_k$ and hence are needed to approximate the expectation in Eq.~\ref{eq:OAT-rollout}. This corresponds to modules 1-3 in in Fig.~\ref{fig:general_overview}.

\subsubsection{Estimating the probability distribution for the number of requests $\tilde{p}_{\eta_{s_k}}$}
We estimate $\tilde{p}_{\eta_{s_k}}$ by leveraging sentence embeddings, spectral clustering and averaging, the demand prediction module, and a novel probabilistic approach that derives a semi-deterministic approximation for $\tilde{p}_{\eta_{s_k}}$ from the predicted demand, obtaining a minute by minute demand distribution from the hourly demand prediction. 

\paragraph{Event data and sentence embeddings}
We need a list of events that includes the event's date, title, description, and venue. Such a list of events can be scraped from the internet using queries \citep{Markou2019} or an online database \citep{PHQ}. Events' attendance depends on multiple factors, including the public's attitude towards such an event. For this reason, we try to estimate a proxy of the public's attitude towards an event by considering reviews for event-venues pairs. We collect $Q$ reviews for each event-venue pair. For each review's textual excerpt we generate a $d$-dimensional sentence embedding using a pre-trained MLM.
Sentence embeddings tend to capture semantic information and inter-word relations in a dense vector representation \citep{cer2018universal, yang-etal-2021-universal}. We denote the $q$-th sentence embedding for a specific event as $\vec{e}_q \in \mathbb{R}^{d}$, and we denote the set of all $Q$ embeddings as $\mathcal{E}$.

\paragraph{Spectral clustering}
For each event, we consider its associated $Q$ review embeddings. We want to obtain semantically relevant representations that aggregate the reviews as dense vectors. To do this we consider spectral clustering on the latent space of the review embeddings. We assume that the latent space for the embeddings is approximately euclidean based on the results presented in \citep{shao2017riemannian}. For our specific application, since we are dealing with an euclidean space, we consider a Gaussian radial basis function as the similarity score $s(\vec{e}_q, \vec{e}_h) = \exp(-\gamma \cdot ||\vec{e}_q - \vec{e}_h||_2^2)$ with $\gamma=1$. To execute spectral clustering, we construct a similarity graph $G_C = (V_C, E_C)$, where each vertex $v_q \in V_C$ corresponds to the review embedding $\vec{e}_q$. We consider a fully connected graph, where all vertices are connected to all other vertices, and the weight of each edge $(v_q, v_h)$ is given by the similarity score $s(\vec{e}_q, \vec{e}_h)$. Using this complete graph, we then apply spectral clustering for $b$ clusters using the algorithm proposed in \citet{Damle2018}. 
We denote the cluster label assigned to embedding $\vec{e}_q$ as $\hat{b}(\vec{e}_q) \in \{ 1, \dots, b\}$. We denote the set of embeddings that have cluster label $a$ for $a \in \{ 1, \dots, b\}$ as $\mathcal{B}_a = \{ \vec{e}_q | \hat{b}(\vec{e}_q)=a, \vec{e}_q \in \mathcal{E} \}$.

\paragraph{Cluster averaging} 
Once all reviews for an event are clustered, we average the embeddings for each assigned cluster to obtain semantically relevant averaged embedding $\hat{e}_{a} = \frac{\sum_{\vec{e}_q\in \mathcal{B}_a} \vec{e}_q}{|\mathcal{B}_a|}$ for each cluster label $a \in \{1, \dots, b\}$. We then stack the resulting $b$ embeddings to obtain a dense representation $R_w = [\hat{e}_{1}^\top, \dots \hat{e}_{b}^\top]^\top$ for arbitrary event $w$. We denote the set of events in sector $s_k$ as $w_{s_k}$. we denote the average dense representation for events in sector $s_k$ as $\hat{R}_{s_k} = \frac{\sum_{w \in w_{s_k}} R_w}{|w_{s_k}|}$. For each event title and description we generate an additional sentence embedding using the same pre-trained MLM. This embedding provides additional context to the NN to differentiate between events that share the same venue and hence might have some of the same reviews. For arbitrary event $w$, We denote this embedding as $\vec{z}_w \in \mathbb{R}^{d}$. We denote the average title embedding for events in sector $s_k$ as $\hat{z}_{s_k} = \frac{\sum_{w \in w_{s_k}} \vec{z}_w}{|w_{s_k}|}$. We define the final unified sector feature for an arbitrary sector $s_k$ as $F_{s_k} = [\hat{z}_{s_k}^\top, \hat{R}_{s_k}^\top]^\top \in \mathbb{R}^{(b+1) \cdot d}$.

\paragraph{Demand prediction module}
The demand prediction module is composed of a temporal (day of the week, month, and hour) and spatial (weather) data collection pipeline followed by a two NN prediction mechanism (see Fig.~\ref{fig:general_overview}.2). If there are no events happening on sector $s_k$, the system considers a NN that takes as input only temporal and spatial data in the form of a vector $\vec{f}$. If there is an event happening on sector $s_k$, the system enhances $\vec{f}$ with the unified sector feature $F_{s_k}$ from the \emph{demand prediction module} to create input feature $F_{s_k}^+ = [\vec{f}^\top, F_{s_k}^\top]^\top$. For simplicity, we denote the predicted demand, irrespective of the NN that produced it, as $\hat{y}_{s_k}$.

\paragraph{Deriving minute demand distribution from hourly predicted demand}
Following the concept of Certainty Equivalence presented in \citet{BertsekasCE}, we derive a semi-deterministic approximation for $\tilde{p}_{\eta_{s_k}}$ by evenly distributing the predicted hourly number of requests for sector $s_k$ over the time horizon $N=60$ to obtain the number of requests that will enter the system at each minute. In this sense, we obtain a single descriptor $\tilde{\eta}_{s_k}^{(\text{det})}$ for all realizations $\tilde{\eta}_{s_k}(t), 1\leq t \leq N$. More formally, We define $\tilde{\eta}_{s_k}^{(\text{det})} = \frac{\hat{y}_{s_k}}{N}$. If $\tilde{\eta}_{s_k}^{(\text{det})} > 1$, we round this value to obtain a deterministic quantity $\bar{\eta}_{s_k}$ that will be used as the number of requests that enter sector $s_k$ at each minute $t$. If $\tilde{\eta}_{s_k}^{(\text{det})} < 1$, then we define a Bernoulli random variable $\tilde{\eta}_{s_k}^\text{bern} \sim Bern(\tilde{\eta}_{s_k}^{(\text{det})})$, and its realization at time $t$ as $\tilde{\eta}_{s_k}^{\text{bern}(t)}$, such that $\tilde{\eta}_{s_k}^{\text{bern}(t)}=1$ with probability $\tilde{\eta}_{s_k}^{(\text{det})}$. We set $\bar{\eta}_{s_k} = \tilde{\eta}_{s_k}^\text{bern}$. For each minute $t$ in time horizon $N=60$, we instantiate $\tilde{\eta}_{s_k}^{\text{bern}(t)}$ to determine if a request appeared in sector $s_k$ at minute $t$. In contrast to standard certainty equivalence \citep{BertsekasCE}, our approach does not fully remove the stochasticity for the random variable $\tilde{\eta}_{s_k}$ to avoid underestimating demand.

\subsubsection{Estimating probability distributions for the pickup and drop-off intersections, $\tilde{p}_{\rho_{s_k}}$ and $\tilde{p}_{\delta_{s_k}}$, and the conditional matching $\tilde{p}_{\beta_{s_k}}$ }
Here, we tackle the problem of going from sector level demand to intersection-level demand that can be used as input to the routing optimization in Eq.\ref{eq:OAT-rollout}. To achieve this, the probability distributions for pickups and drop-offs over intersections, $\tilde{p}_{\rho_{s_k}}$ and $\tilde{p}_{\delta_{s_k}}$, are estimated by leveraging the demand assignment module (see Fig.~\ref{fig:general_overview}.3), which given a set of intersections $\mathcal{I}_{s_k}$ for sector $s_k$, returns a probability distribution over $\mathcal{I}_{s_k}$ based on the number of locales of interest (i.e. restaurants, bars, cafes, etc) near that intersection, their respective occupancy schedule \citep{Happle2019, PANCHABIKESAN2021119539}, and an estimate of their maximum occupancy.
\begin{figure}
    \centering
    \includegraphics[width=0.7\linewidth]{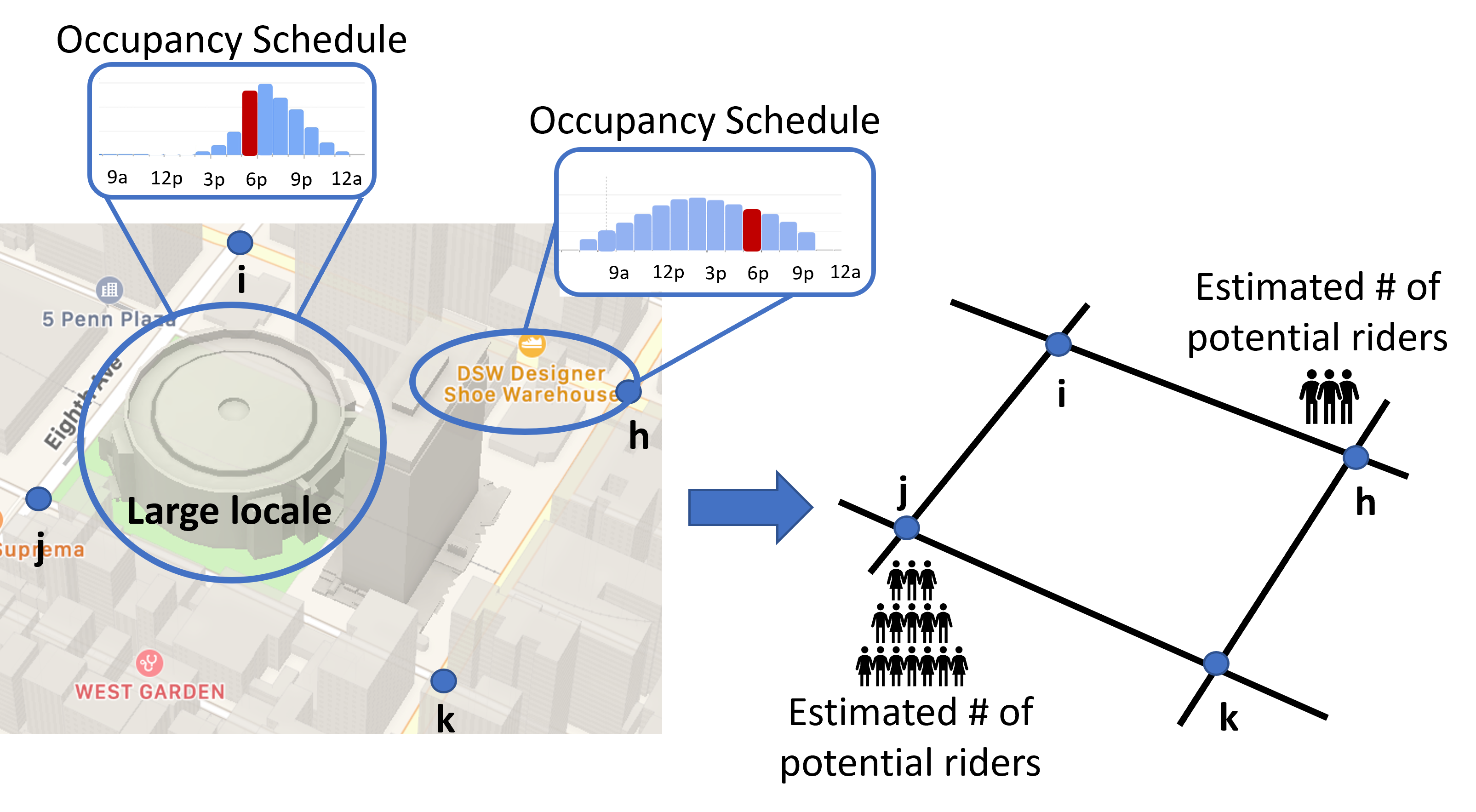}
    \vspace{-10pt}
    \caption{\small{Mapping probability distributions over sectors to intersections for pickup and drop-off. The potential number of riders at each intersection is estimated using occupancy schedules and the maximum occupancy of the locale}}
    \Description{Graphical representation of our proposed procedure for mapping probability distributions over sectors to intersections for pickup and drop-off. The potential number of riders at each intersection is estimated using occupancy schedules and the maximum occupancy of the locale}
    \label{fig:occupancy_schedules}
\end{figure}
A graphical representation of this process is shown in Fig.~\ref{fig:occupancy_schedules}. More formally, let's consider sector $s_k$ and an intersection $j$ such that $j \in \mathcal{I}_{s_k}$. We denote $n_{\lambda}(j)$, as the number of locales for a specific locale type $\lambda \in \Lambda$ near intersection $j$, where $\Lambda$ is the set of all locale types. We denote $o_{\lambda}$ and $p_{\lambda}$, as the estimated maximum occupancy and the occupancy schedule (percentage of maximum occupancy) for locale type $\lambda$, respectively. For a given hour $h_t$, we denote the percentage occupancy of locale type $\lambda$ as $p_{\lambda}(h_t)$. We denote the estimate of the number of potential customers at node $j$ given an hour $h_t$ as $O(j, h_t) = \sum_{\lambda \in \Lambda} n_{\lambda}(j) \cdot (p_{\lambda}(h_t) \cdot o_{\lambda})$. To obtain an estimate of $\tilde{p}_{\rho_{s_k}}$ and $\tilde{p}_{\delta_{s_k}}$ at a given hour $h_t$, we normalize $O(j, h_t)$ for each node $j$ by dividing it by $\sum_{i \in \mathcal{I}_{s_k}} O(i, h_t)$.

To estimate $\tilde{p}_{\beta_{s_k}}$, the probability distribution of having a request being dropped off at any sector given that it was picked up at sector $s_k$, we consider historical demand data. We estimate the conditional probability distribution by looking at the relative frequency of pickup-drop-off sector pairs in the data. 

\section{Experimental evaluation}
\label{subsec:setup}
To evaluate our approach, we consider ride-share trip data for the year 2022 obtained from NYC's HV-FHV datasets \cite{TLC_Data}. We consider a section of the map with 38 sectors (2235 intersections and 4566 streets) across mid and lower Manhattan (see Fig.~\ref{fig:example_env}). We collect data for more than $2,000$ different events during the year 2022 hosted on 300 unique venues across all 38 sectors. We consider that each time step in the system corresponds to $1$ minute, and the time horizon $N=60$ represents an hour.

\subsection{Implementation details}
To implement the system, we need to deal with two main areas:

\subsubsection{Review data and spectral clustering:} 
We obtain a list of events for a specific date using the PredictHQ API \citep{PHQ}. We consider seven event categories: conferences, expos, concerts, festivals, performing arts, community gatherings, and sports events. For each of these events, we scrape Google Maps using SerpAPI \cite{SerpAPI} to obtain reviews related to an event and its venue. We set $K=100$ to collect a maximum of $100$ reviews. We choose RoBERTa large v1 \cite{Liu2019RoBERTaAR} as the MLM for sentence embeddings, 
We choose the number of clusters for the review embeddings $b=3$ based on previous work on sentiment analysis \citep{Karfi2018, Zimbra2018} that suggests $3$ clusters (positive, negative and neutral reviews). However, $b$ can be set to any number, as long as it is large enough to capture the semantic diversity of the reviews, but still small enough to produce reasonable sized input features for the neural networks. 

\subsubsection{Demand prediction, demand assignment and model-based RL routing modules:}
All the weather data is obtained using Open-Meteo Historical Weather Data API\citep{Open-Meteo}, which uses ERA5\citep{ERA5, ERA5-land}. We train both NNs for $100$ epochs using the first $8$ months of data for the year $2022$. The remaining $4$ months were held out as a test set. For both NNs, we select mean squared error (MSE) as the loss function, we use Adam with a learning rate of $1\times 10^{-4}$ and a weight decay of $1\times 10^{-6}$ as an optimizer, we choose a batch size of 64, and we shuffle the batches after every epoch. The NN that predicts demand using $\vec{f}$ as input has 2 fully-connected hidden layers, each with 256 neurons. The NN that predicts demand using the features $F_{s_k}^{+}$ has 2 fully-connected hidden layers with 4096 neurons each. Architectural parameters for both NNs were selected using a hyper-parameter search, and a simple feed forward architecture was chosen as it was the fastest architecture to train, while still obtaining comparable results to other more complex architectures. All training was done on a single NVIDIA RTX A6000. All evaluation results are calculated using the predictions for the 4 months of data in the test set. For the demand assignment module, we retrieve all locales of interest for each intersection $j \in \mathcal{I}_{V}$ using Google Maps' Nearby Search API \citep{Google_Maps}. We consider four locale types: retail spaces, restaurants, hotels, and hospitals. We retrieve occupancy schedules from COMNET \citep{Comnet2016}. For the rollout-based routing module, we set the planning horizon $H = 10$ and use $1000$ Monte-Carlo simulations to estimate the expectation for the one-agent-at-a-time online minimization, and choose $\bar{\pi}$ to be the fastest computing heuristic, the greedy policy (see Sec.~\ref{subsec:baselines}). It is important to note that our proposed method will still work for base policies that are easy to compute and have a reasonable behavior.

\subsection{Baselines}
\label{subsec:baselines}
Our main results compare our approach against the following four baselines:
\textbf{1) Greedy policy:} available taxis are routed to their nearest outstanding request as given by Dijkstra's shortest path algorithm without coordination among them. 
\textbf{2) Instantaneous assignment:}uses a variation of broadcast of local eligibility (BLE)~\citep{werger2000broadcast} for iterative task assignment, performing a deterministic instantaneous matching of available taxis to request currently in the system. 
\textbf{3) \citet{garces2023}:} in its scalable implementation, this methods uses a one-agent-a-time rollout with instantaneous assignment as the base policy. It estimates the current demand using the demand of the previous hour of operation of the system. This method uses a standard application of certainty equivalence, utilizing the mean values for $\tilde{\eta}_{s_h}, \tilde{\rho}_{s_h}, \tilde{\delta}_{s_h}$, but preserving the stochasticity for $\tilde{\beta}_{s_h}$ and the order in which requests arrive by independently sampling from a pre-computed pool of pickups and drop-offs. 
\textbf{4) Oracle:} This method has full a-priori knowledge of the exact time, pickup, drop-off locations for the requests entering the system in the future. For this reason, this method is able to route taxis to the exact location of each request even before the request has entered the system. This method minimizes the total wait time for all requests by executing a series of assignments that leverage the auction algorithm \citep{Bertsekas1979Auction} with full future information. This method is not achievable in practice, but it provides a lower bound on the cost.

Note that we do not compare against policy gradient methods as applying these methods to a large scale multiagent routing environment with demand surges is still an open problem.

\subsection{Main results}
In this section, we present a comparative study of our proposed approach and the baselines described in Sec.~\ref{subsec:baselines}. We evaluate all policies using the held out test dataset described in Sec.~\ref{subsec:setup}. We use the real requests in the ride service data \citep{TLC_Data} as the ground truth requests entering the system, and we select a time window from 3pm to 9pm to include pre-surge scenarios (3pm), event-driven surge times (5pm and 7pm), and post-surge times (9pm). In the selected time window events happen 4 to 6 days a week usually in the selected time range (start time of concerts and theater shows). All results are obtained by averaging results for $25$ random initial states for a randomly chosen date (in this case, November 17, 2022).

First, we compare the average total cost (total wait time for all requests at the end of the horizon) incurred by each policy. We present these results in Table~\ref{tab:total_cost}. As shown in Table~\ref{tab:total_cost}, our proposed approach obtains the lowest average total cost compared to all the other feasible methods, having a $2\%$ to $10\%$ lower cost depending on the hour considered. Compared to state-of-the-art rollout-based routing \citet{garces2023}, which incurs a high cost during demand surges, our method is able to use event information to update the demand model and successfully anticipate event-driven demand surges, maintaining low costs.

\begin{table*}[ht]
    \caption{ \small Average total cost of different policies}
    \label{tab:total_cost}
  \centering
  \begin{tabular}{lllll|l}
    \toprule
    & \multicolumn{5}{c}{Policies}                   \\
    \cmidrule(r){2-6}
    Time of day & Greedy & Inst. Assign. & Garces et al. & Our approach & Oracle  \\
    \midrule
    3pm & $6671.2 \pm 269.0$ & $6670.6 \pm 193.9$ & $6563.9 \pm 243.5$ & $\mathbf{6121.5 \pm 217.5}$ & $5732.1 \pm 165.4$ \\
    5pm & $11795.7 \pm 202.6$ & $11848.0 \pm 165.1$ & $11667.9 \pm 174.5$ & $\mathbf{11291.7 \pm 169.2}$ & $11082.0 \pm 134.7$\\
    7pm & $17199.8 \pm 187.7$ & $17432.2 \pm 187.0$ & $18070.6 \pm 210.6$ & $\mathbf{16868.2 \pm 141.2}$ & $16573.4 \pm 103.8$ \\
    9pm & $12102.4 \pm 143.1$ & $12460.4 \pm 107.4$ & $13147.4 \pm 217.5$ & $\mathbf{12020.6 \pm 154.6}$ & $11552.6 \pm 80.7$\\
    \bottomrule
  \end{tabular}
\end{table*}

\begin{figure}
    \centering
    \includegraphics[width=0.58\linewidth]{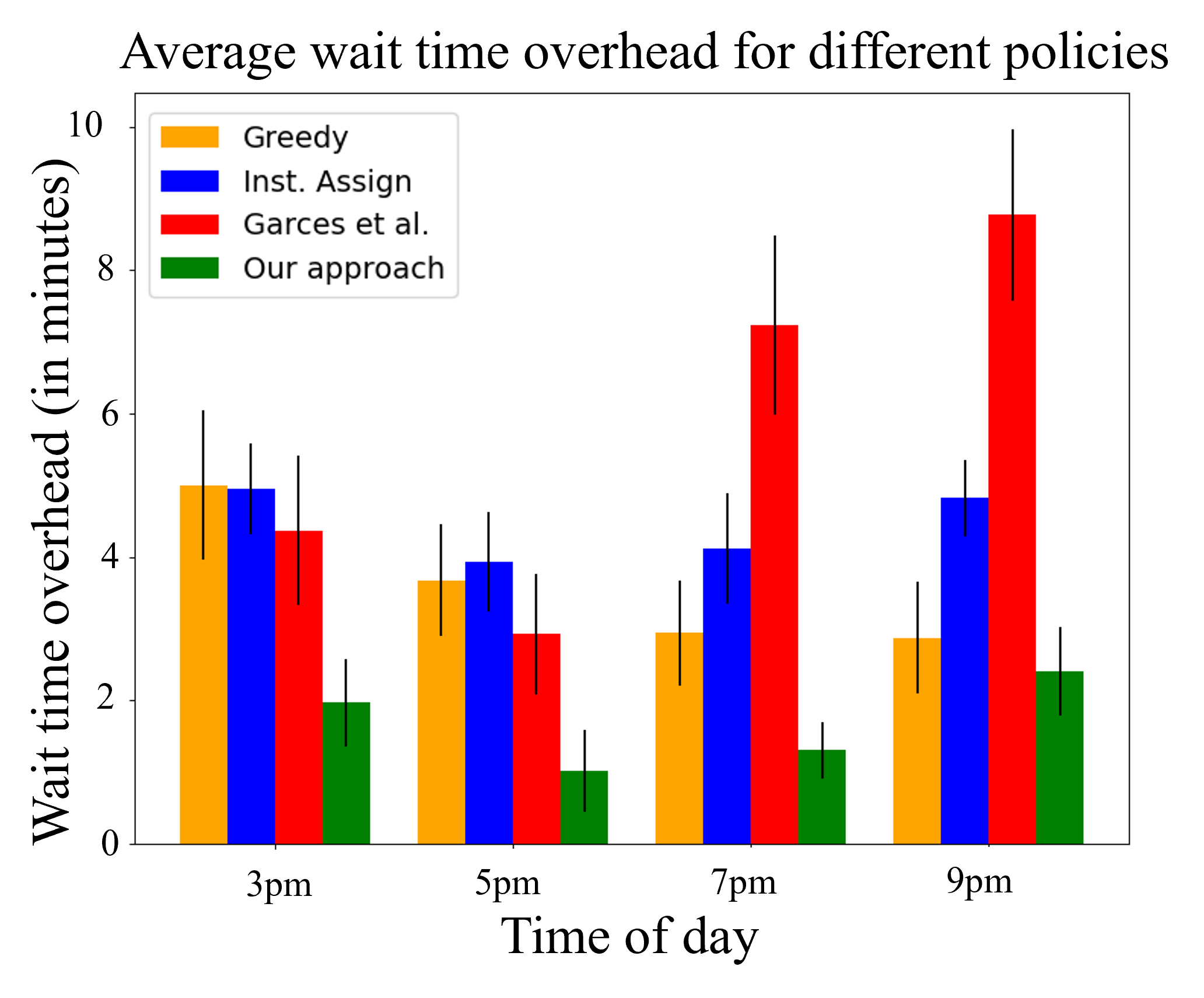}
    \vspace{-10pt}
    \caption{\small{Average wait time overhead per serviced request (in minutes)}}
    \label{fig:user_wait_time}
    \Description{Bar graph showing the average wait time overhead per serviced request (in minutes) for different hour of the daya for our approach and all the other baselines.}
    \vspace{-10pt}
\end{figure}

We also present results for the average wait time overhead per serviced request in order to better understand the impact of each policy on rider experience. Intuitively wait time overhead is the additional amount of time that a request will have to wait in a realistic stochastic setting compared to a setting in which the locations and times of all requests are known a-priori. To calculate the wait time overhead per serviced request we take the total wait time for all requests for a given policy, subtract the total wait time for the oracle and then divide the resulting value by the number of serviced requests for the given policy. We present these results in Fig.~\ref{fig:user_wait_time}. As shown in Fig.~\ref{fig:user_wait_time}, our approach results in $25\%$ to $75\%$ improvement on wait time overhead per serviced request over all the other methods. Our method is particularly useful in surge times between 5pm and 7pm, as it results in wait time overheads that are 3X smaller than wait time overheads for all the other methods.

Additionally, we present the number of outstanding requests at the end of the horizon to see which feasible policy services more requests. Since we are dealing with a fixed fleet-size that is chosen to satisfy average demand scenarios, demand surges usually lead to accumulation of unserviced requests at the end of the horizon. Table~\ref{tab:outstanding_requests} contains the results for the number of outstanding requests at the end of the horizon for all methods. As shown in Table~\ref{tab:outstanding_requests}, our method has $1\%$ to $4\%$ fewer outstanding requests at the end of the horizon. Since the objective of both our method and the oracle is to minimize the total wait time for all requests, picking up earlier longer requests can result in the smallest total wait, specially during surge times. Servicing longer requests means that the taxis would be occupied for longer and hence cannot pick up other requests. However, since these longer requests enter the system earlier than other requests, not picking them up will have a large impact on the total cost. Due to the stochasticity of the demand model for our method and the large volume of requests entering the system during surge times, our method is not able to capture the benefits of these earlier longer requests giving priority to shorter trips whose cost can be accurately estimated using the planning horizon of $10$ minutes. For this reason, we see that the oracle produces a lower cost but services fewer requests than our method during surge times (5pm  and 7pm).

\begin{table*}[ht]
  \caption{ \small Average number of outstanding requests at the end of the hour for different policies}
  \label{tab:outstanding_requests}
  \centering
  \begin{tabular}{lllll|l}
    \toprule
    & \multicolumn{5}{c}{Policies}                   \\
    \cmidrule(r){2-6}
    Time of day & Greedy & Inst. Assign. & Garces et al. & Our approach & Oracle \\
    \midrule
    3pm & $241.92 \pm 7.87$ & $240.92 \pm 3.75$ & $238.42 \pm 4.66$ & $\mathbf{230.17 \pm 4.30}$ & $230.25 \pm 3.25$\\
    5pm & $434.00 \pm 4.07$ & $433.33 \pm 3.52$ & $426.83 \pm 4.76$ & $\mathbf{422.58 \pm 4.25}$ & $426.75 \pm 3.24$\\
    7pm & $626.23 \pm 5.15$ & $630.15 \pm 5.49$ & $632.38 \pm 4.89$ & $\mathbf{611.46 \pm 3.71}$ & $620.92 \pm 2.81$\\
    9pm & $471.14 \pm 5.71$ & $474.93 \pm 3.63$ & $479.79 \pm 3.53$ & $\mathbf{467.14 \pm 5.04}$ & $465.36 \pm 2.78$\\
    \bottomrule
  \end{tabular}
\end{table*}

From all these results, we obtain that our method does not only decreases wait time overhead for serviced requests by $25\%$ to $75\%$, depending on the hour, but it also services $1\%$ to $4\%$ more requests than all the other methods.

\begin{wraptable}[9]{l}{0.5\linewidth}
  \centering
  \vspace{-5pt}
  \caption{\small Average computation time for planning for a single time step (in seconds)}
  \vspace{-5pt}
  \label{tab:runtime_results}
  \resizebox{\linewidth}{!}{
  \begin{tabular}{ll}
    \toprule
    Policy & Runtime (in seconds)\\
    \midrule
     Garces et al. & $153.42 \pm 7.65$ \\
     Our approach & $56.47 \pm 5.40$\\
    \bottomrule
  \end{tabular}}
\end{wraptable}

To emphasize the computational time reductions associated with our approach compared to standard one-agent-at-a-time rollout, we present computation times for the method in \citet{garces2023} (see Sec.~\ref{subsec:baselines}) and for our approach in Table~\ref{tab:runtime_results}. As shown in Table~\ref{tab:runtime_results} our method is able to plan for the next time step in approximately a third of the time compared to \citet{garces2023}, making it more suitable for city-scale applications.

\subsection{Ablation studies}
\label{subsec:ablation_studies}
In this section we consider modifications of our approach, mainly dealing with two aspects: the demand prediction system and the base policy used for the rollout-based routing algorithm.

\subsubsection{Demand prediction system}
\begin{figure}
    \centering
    \includegraphics[width=0.55\linewidth]{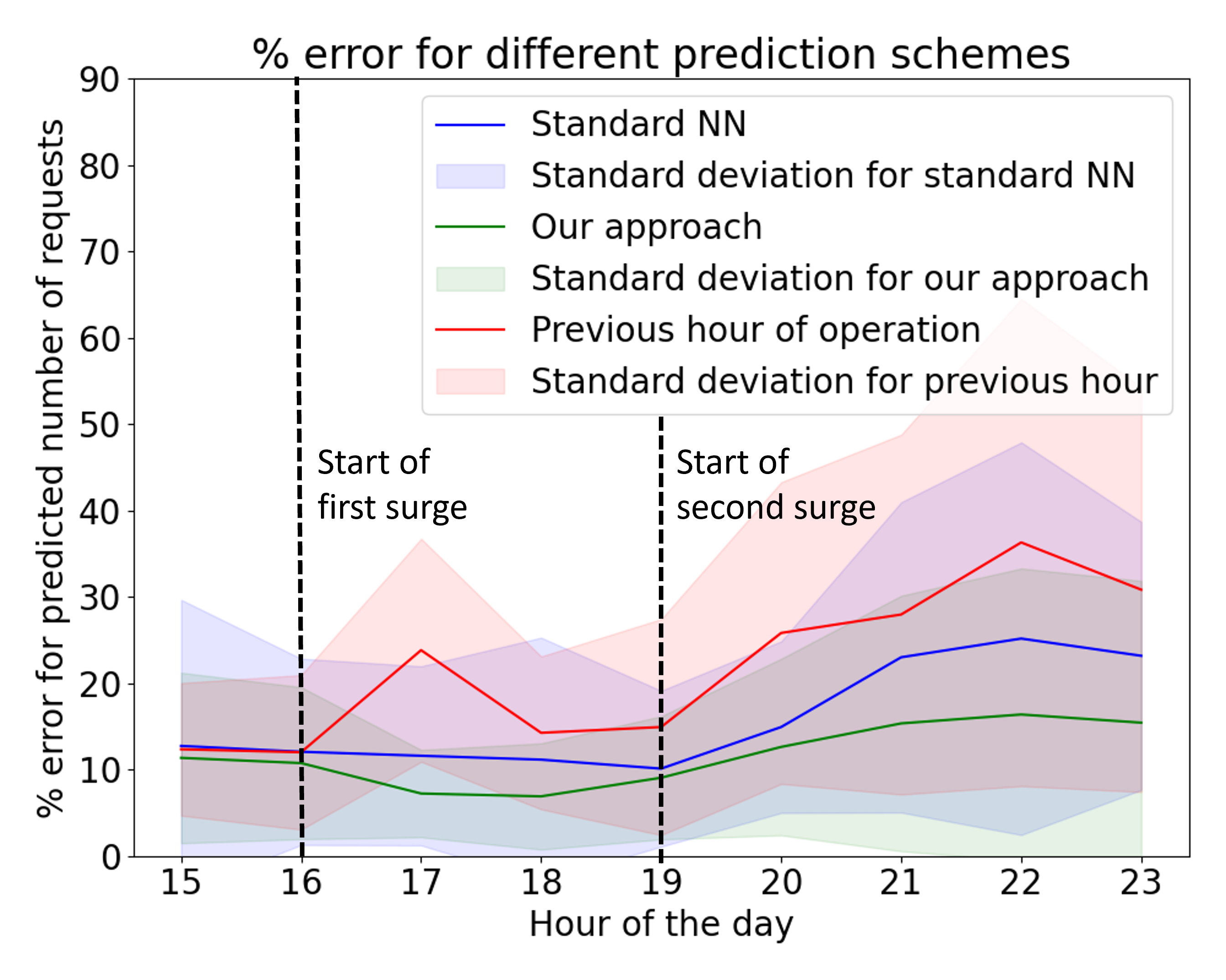}
    \vspace{-10pt}
    \caption{\small{$PE_{s_k}$ for different demand prediction schemes}}
    \Description{Line graph showing the prediction error for our approach compared to a neural network that only uses spatial and temporal data, and a probabilistic approach that uses demand during the previous hour of operation to estimate the current demand model.}
    \label{fig:monthly_predictive_error}
\end{figure}

To better understand the effect of including event information in the demand prediction, we isolate the demand prediction system and compare it against a NN that only uses temporal and spatial data as input (we call this standard NN). We also compare our demand prediction against a probabilistic estimation of demand that uses the previous hour of operation of the system as a proxy for the current demand as proposed in \citep{garces2023}. Results for the prediction errors of all three methods are shown in Fig.~\ref{fig:monthly_predictive_error}.The figure presents prediction errors for each hour in the chosen time window averaged over all days in November 2022, where we compute average percent error $PE_{s_k} = \frac{1}{|X|} \sum_{X} \frac{|y_{s_k} - \hat{y}_{s_k}|}{y_{s_k}}$, where $X$ is the number of data points in the test set for sector $s_k$ at the hour of interest, $\hat{y}_{s_k}$ is the predicted number of requests entering sector $s_k$, and $y_{s_k}$ is the actual number of requests entering sector $s_k$.  Our prediction scheme outperforms all other methods, obtaining $3\%-10\%$ improvement on average percent error.


\subsubsection{Base policy}
\begin{wraptable}{l}{0.5\linewidth}
  \centering
  \vspace{-10pt}
    \caption{\small Average total cost of our approach for different base policies}
  \label{tab:ablation_base_policy}
  \vspace{-5pt}
  \centering
  \resizebox{\linewidth}{!}{
  \begin{tabular}{lll}
    \toprule
    & \multicolumn{2}{c}{Base Policies}                   \\
    \cmidrule(r){2-3}
    Time of day & Inst. Assign. & Greedy  \\
    \midrule
    3pm & $103.73 \pm 3.32$ & $102.57 \pm 3.34$  \\
    5pm & $189.25 \pm 3.23$ & $189.03 \pm 2.12$ \\
    7pm & $279.37 \pm 2.47$ & $280.88 \pm 2.79$  \\
    9pm & $202.84 \pm 2.36$ & $201.79 \pm 2.06$ \\
    \bottomrule
  \end{tabular}}
\end{wraptable}
Our proposed approach does not depend on the choice of the base policy as long as the base policy is reasonable and easy to compute. To illustrate this point, we consider an ablation study where we compare the performance of our approach when we change the base policy. We consider instantaneous assignment and a greedy policy as defined in Sec.~\ref{subsec:baselines} as potential candidates for base policies. The results of this comparison are shown in Table~\ref{tab:ablation_base_policy}, where we can see that the total cost of our method at the end of the time horizon is similar for both base policies.


\section{Limitations and future work}
Since we consider a fixed fleet size, larger demand surges still lead to larger number of outstanding requests compared to hours where there are less events. As future work, we want to consider a system where there are taxis stored at warehouses throughout the map, and we can dynamically change the fleet size to address this rise in demand. Another important limitation of our approach is that it relies on sampling to estimate the expected cost of each action, and hence there is still a small, but non-zero, probability that the samples obtained might not be representative of the real demand, leading to a degrade in performance. As shown in the simulations, this probability is very small and performance degradation happens very infrequently.

\section{Conclusion}
In this paper we presented an event-informed multi-agent RL routing framework that leverages event data from the internet to predict demand surges and route taxis accordingly. The framework proposed in this paper is applicable to multiple multiagent sequential decision making problems where unstructured textual data can be used to predict changes in the environment. One of such problems is the routing application covered in the paper, in which event data (textual data) is leveraged to predict changes in the demand (the environment). Other examples of potential applications are: multirobot repair/maintenance problems where there are textual descriptions of the conditions of facilities, robotic search-and-rescue tasks where humans collaborate with robots, among others. In the routing application, our method decreases wait time overhead for serviced requests by $25\%$ to $75\%$, while servicing $1\%$ to $4\%$ more requests than all the other methods.


\begin{acks}
The authors gratefully acknowledge ONR award $\#$N00014-21-1-2714, AFOSR award $\#$FA9550-22-1-0223 and NSF CAREER award $\#$2114733 for partial support of this work.
\end{acks}



\bibliographystyle{ACM-Reference-Format} 
\bibliography{refs}


\begin{thebibliography}{62}


\ifx \showCODEN    \undefined \def \showCODEN     #1{\unskip}     \fi
\ifx \showDOI      \undefined \def \showDOI       #1{#1}\fi
\ifx \showISBNx    \undefined \def \showISBNx     #1{\unskip}     \fi
\ifx \showISBNxiii \undefined \def \showISBNxiii  #1{\unskip}     \fi
\ifx \showISSN     \undefined \def \showISSN      #1{\unskip}     \fi
\ifx \showLCCN     \undefined \def \showLCCN      #1{\unskip}     \fi
\ifx \shownote     \undefined \def \shownote      #1{#1}          \fi
\ifx \showarticletitle \undefined \def \showarticletitle #1{#1}   \fi
\ifx \showURL      \undefined \def \showURL       {\relax}        \fi
\providecommand\bibfield[2]{#2}
\providecommand\bibinfo[2]{#2}
\providecommand\natexlab[1]{#1}
\providecommand\showeprint[2][]{arXiv:#2}

\bibitem[\protect\citeauthoryear{Ahamed, Zou, Farazi, and Tulabandhula}{Ahamed et~al\mbox{.}}{2021}]%
        {AHAMED2021227}
\bibfield{author}{\bibinfo{person}{Tanvir Ahamed}, \bibinfo{person}{Bo Zou}, \bibinfo{person}{Nahid~Parvez Farazi}, {and} \bibinfo{person}{Theja Tulabandhula}.} \bibinfo{year}{2021}\natexlab{}.
\newblock \showarticletitle{Deep Reinforcement Learning for Crowdsourced Urban Delivery}.
\newblock \bibinfo{journal}{\emph{Transportation Research Part B: Methodological}}  \bibinfo{volume}{152} (\bibinfo{year}{2021}), \bibinfo{pages}{227--257}.
\newblock
\showISSN{0191-2615}
\urldef\tempurl%
\url{https://doi.org/10.1016/j.trb.2021.08.015}
\showDOI{\tempurl}


\bibitem[\protect\citeauthoryear{ASHRAE}{ASHRAE}{2013}]%
        {ASHRAE2013}
\bibfield{author}{\bibinfo{person}{ASHRAE}.} \bibinfo{year}{2013}\natexlab{}.
\newblock \bibinfo{title}{ANSI/ASHRAE/IES Standard 90.1-2013: Energy Standard for Buildings Except Low-Rise Residential Buildings}.
\newblock
\newblock


\bibitem[\protect\citeauthoryear{Barkan, Razin, Malkiel, Katz, Caciularu, and Koenigstein}{Barkan et~al\mbox{.}}{2019}]%
        {barkan2019scalable}
\bibfield{author}{\bibinfo{person}{Oren Barkan}, \bibinfo{person}{Noam Razin}, \bibinfo{person}{Itzik Malkiel}, \bibinfo{person}{Ori Katz}, \bibinfo{person}{Avi Caciularu}, {and} \bibinfo{person}{Noam Koenigstein}.} \bibinfo{year}{2019}\natexlab{}.
\newblock \bibinfo{title}{Scalable Attentive Sentence-Pair Modeling via Distilled Sentence Embedding}.
\newblock
\newblock
\showeprint[arxiv]{1908.05161}~[cs.LG]


\bibitem[\protect\citeauthoryear{Bent and Van~Hentenryck}{Bent and Van~Hentenryck}{2004}]%
        {Bent2004a}
\bibfield{author}{\bibinfo{person}{Russell Bent} {and} \bibinfo{person}{Pascal Van~Hentenryck}.} \bibinfo{year}{2004}\natexlab{}.
\newblock \showarticletitle{The Value of Consensus in Online Stochastic Scheduling.}. In \bibinfo{booktitle}{\emph{Proceedings of the 14th International Conference on Automated Planning and Scheduling, ICAPS 2004}}. \bibinfo{publisher}{American Association for Artificial Intelligence}, \bibinfo{address}{Providence, RI}, \bibinfo{pages}{219--226}.
\newblock


\bibitem[\protect\citeauthoryear{Bertsekas}{Bertsekas}{1979}]%
        {Bertsekas1979Auction}
\bibfield{author}{\bibinfo{person}{D.P. Bertsekas}.} \bibinfo{year}{1979}\natexlab{}.
\newblock \showarticletitle{A Distributed Algorithm for the Assignment Problem}.
\newblock \bibinfo{journal}{\emph{Lab for Information and Decision Systems Report}} (\bibinfo{date}{05} \bibinfo{year}{1979}).
\newblock


\bibitem[\protect\citeauthoryear{Bertsekas}{Bertsekas}{2020a}]%
        {BERTSEKAS2020Multiagent}
\bibfield{author}{\bibinfo{person}{Dimitri Bertsekas}.} \bibinfo{year}{2020}\natexlab{a}.
\newblock \showarticletitle{Multiagent value iteration algorithms in dynamic programming and reinforcement learning}.
\newblock \bibinfo{journal}{\emph{Results in Control and Optimization}}  \bibinfo{volume}{1} (\bibinfo{year}{2020}), \bibinfo{pages}{100003}.
\newblock
\showISSN{2666-7207}
\urldef\tempurl%
\url{https://doi.org/10.1016/j.rico.2020.100003}
\showDOI{\tempurl}


\bibitem[\protect\citeauthoryear{Bertsekas}{Bertsekas}{2020b}]%
        {bertsekas2020rollout}
\bibfield{author}{\bibinfo{person}{D. Bertsekas}.} \bibinfo{year}{2020}\natexlab{b}.
\newblock \bibinfo{booktitle}{\emph{Rollout, Policy Iteration, and Distributed Reinforcement Learning}}.
\newblock \bibinfo{publisher}{Athena Scientific.}
\newblock
\showISBNx{9781886529076}
\urldef\tempurl%
\url{https://books.google.com/books?id=Hbo-EAAAQBAJ}
\showURL{%
\tempurl}


\bibitem[\protect\citeauthoryear{Bertsekas}{Bertsekas}{2021}]%
        {Bertsekas2021PI}
\bibfield{author}{\bibinfo{person}{Dimitri Bertsekas}.} \bibinfo{year}{2021}\natexlab{}.
\newblock \showarticletitle{Multiagent Reinforcement Learning: Rollout and Policy Iteration}.
\newblock \bibinfo{journal}{\emph{IEEE/CAA Journal of Automatica Sinica}} \bibinfo{volume}{8}, \bibinfo{number}{2} (\bibinfo{year}{2021}), \bibinfo{pages}{249--272}.
\newblock
\urldef\tempurl%
\url{https://doi.org/10.1109/JAS.2021.1003814}
\showDOI{\tempurl}


\bibitem[\protect\citeauthoryear{Bertsekas}{Bertsekas}{2022}]%
        {Bertsekas2022AlphaZero}
\bibfield{author}{\bibinfo{person}{Dimitri Bertsekas}.} \bibinfo{year}{2022}\natexlab{}.
\newblock \bibinfo{booktitle}{\emph{Lessons from AlphaZero for Optimal, Model Predictive, and Adaptive Control}}.
\newblock \bibinfo{publisher}{Athena Scientific}, \bibinfo{address}{Nashua, NH, USA}.
\newblock
\showISBNx{1886529175}


\bibitem[\protect\citeauthoryear{Bertsekas and Castanon}{Bertsekas and Castanon}{1998}]%
        {BertsekasCE}
\bibfield{author}{\bibinfo{person}{D.P. Bertsekas} {and} \bibinfo{person}{D.A. Castanon}.} \bibinfo{year}{1998}\natexlab{}.
\newblock \showarticletitle{Rollout algorithms for stochastic scheduling problems}. In \bibinfo{booktitle}{\emph{Proceedings of the 37th IEEE Conference on Decision and Control (Cat. No.98CH36171)}}, Vol.~\bibinfo{volume}{2}. \bibinfo{pages}{2143--2148 vol.2}.
\newblock
\urldef\tempurl%
\url{https://doi.org/10.1109/CDC.1998.758655}
\showDOI{\tempurl}


\bibitem[\protect\citeauthoryear{Bertsimas, Jaillet, and Martin}{Bertsimas et~al\mbox{.}}{2019}]%
        {Bertsimas2019OnlineVR}
\bibfield{author}{\bibinfo{person}{Dimitris Bertsimas}, \bibinfo{person}{Patrick Jaillet}, {and} \bibinfo{person}{S{\'e}bastien Martin}.} \bibinfo{year}{2019}\natexlab{}.
\newblock \showarticletitle{Online Vehicle Routing: The Edge of Optimization in Large-Scale Applications}.
\newblock \bibinfo{journal}{\emph{Oper. Res.}}  \bibinfo{volume}{67} (\bibinfo{year}{2019}), \bibinfo{pages}{143--162}.
\newblock


\bibitem[\protect\citeauthoryear{Bhattacharya, Kailas, Badyal, Gil, and Bertsekas}{Bhattacharya et~al\mbox{.}}{2021}]%
        {Bhattacharya2020MultiagentRollout}
\bibfield{author}{\bibinfo{person}{Sushmita Bhattacharya}, \bibinfo{person}{Siva Kailas}, \bibinfo{person}{Sahil Badyal}, \bibinfo{person}{Stephanie Gil}, {and} \bibinfo{person}{Dimitri Bertsekas}.} \bibinfo{year}{2021}\natexlab{}.
\newblock \showarticletitle{Multiagent Rollout and Policy Iteration for POMDP with Application to Multi-Robot Repair Problems}. In \bibinfo{booktitle}{\emph{Proceedings of the 2020 Conference on Robot Learning}} \emph{(\bibinfo{series}{Proceedings of Machine Learning Research}, Vol.~\bibinfo{volume}{155})}, \bibfield{editor}{\bibinfo{person}{Jens Kober}, \bibinfo{person}{Fabio Ramos}, {and} \bibinfo{person}{Claire Tomlin}} (Eds.). \bibinfo{publisher}{PMLR}, \bibinfo{pages}{1814--1828}.
\newblock
\urldef\tempurl%
\url{https://proceedings.mlr.press/v155/bhattacharya21a.html}
\showURL{%
\tempurl}


\bibitem[\protect\citeauthoryear{Cao, Zeng, Wang, Sharma, Ma, Liu, and Zhou}{Cao et~al\mbox{.}}{2022}]%
        {Cao2022}
\bibfield{author}{\bibinfo{person}{Dun Cao}, \bibinfo{person}{Kai Zeng}, \bibinfo{person}{Jin Wang}, \bibinfo{person}{Pradip~Kumar Sharma}, \bibinfo{person}{Xiaomin Ma}, \bibinfo{person}{Yonghe Liu}, {and} \bibinfo{person}{Siyuan Zhou}.} \bibinfo{year}{2022}\natexlab{}.
\newblock \showarticletitle{BERT-Based Deep Spatial-Temporal Network for Taxi Demand Prediction}.
\newblock \bibinfo{journal}{\emph{IEEE Transactions on Intelligent Transportation Systems}} \bibinfo{volume}{23}, \bibinfo{number}{7} (\bibinfo{year}{2022}), \bibinfo{pages}{9442--9454}.
\newblock
\urldef\tempurl%
\url{https://doi.org/10.1109/TITS.2021.3122114}
\showDOI{\tempurl}


\bibitem[\protect\citeauthoryear{Cer, Yang, yi~Kong, Hua, Limtiaco, John, Constant, Guajardo-Cespedes, Yuan, Tar, Sung, Strope, and Kurzweil}{Cer et~al\mbox{.}}{2018}]%
        {cer2018universal}
\bibfield{author}{\bibinfo{person}{Daniel Cer}, \bibinfo{person}{Yinfei Yang}, \bibinfo{person}{Sheng yi Kong}, \bibinfo{person}{Nan Hua}, \bibinfo{person}{Nicole Limtiaco}, \bibinfo{person}{Rhomni~St. John}, \bibinfo{person}{Noah Constant}, \bibinfo{person}{Mario Guajardo-Cespedes}, \bibinfo{person}{Steve Yuan}, \bibinfo{person}{Chris Tar}, \bibinfo{person}{Yun-Hsuan Sung}, \bibinfo{person}{Brian Strope}, {and} \bibinfo{person}{Ray Kurzweil}.} \bibinfo{year}{2018}\natexlab{}.
\newblock \bibinfo{title}{Universal Sentence Encoder}.
\newblock
\newblock
\showeprint[arxiv]{1803.11175}~[cs.CL]


\bibitem[\protect\citeauthoryear{Chen, Thakuriah, and Ampountolas}{Chen et~al\mbox{.}}{2021}]%
        {Chen2021UberNet}
\bibfield{author}{\bibinfo{person}{Long Chen}, \bibinfo{person}{Piyushimita Thakuriah}, {and} \bibinfo{person}{Konstantinos Ampountolas}.} \bibinfo{year}{2021}\natexlab{}.
\newblock \showarticletitle{Short-Term Prediction of Demand for Ride-Hailing Services: A Deep Learning Approach}.
\newblock \bibinfo{journal}{\emph{Journal of Big Data Analytics in Transportation}}  \bibinfo{volume}{3} (\bibinfo{date}{08} \bibinfo{year}{2021}).
\newblock
\urldef\tempurl%
\url{https://doi.org/10.1007/s42421-021-00041-4}
\showDOI{\tempurl}


\bibitem[\protect\citeauthoryear{Chen, Wang, and Meng}{Chen et~al\mbox{.}}{2020}]%
        {Chen2020}
\bibfield{author}{\bibinfo{person}{Shukai Chen}, \bibinfo{person}{Hua Wang}, {and} \bibinfo{person}{Qiang Meng}.} \bibinfo{year}{2020}\natexlab{}.
\newblock \showarticletitle{Solving the first-mile ridesharing problem using autonomous vehicles}.
\newblock \bibinfo{journal}{\emph{Computer-Aided Civil and Infrastructure Engineering}} \bibinfo{volume}{35}, \bibinfo{number}{1} (\bibinfo{year}{2020}), \bibinfo{pages}{45--60}.
\newblock
\urldef\tempurl%
\url{https://doi.org/10.1111/mice.12461}
\showDOI{\tempurl}
\showeprint{https://onlinelibrary.wiley.com/doi/pdf/10.1111/mice.12461}


\bibitem[\protect\citeauthoryear{Chng, Anowar, and Cheah}{Chng et~al\mbox{.}}{2022}]%
        {Chng2022}
\bibfield{author}{\bibinfo{person}{Samuel Chng}, \bibinfo{person}{Sabreena Anowar}, {and} \bibinfo{person}{Lynette Cheah}.} \bibinfo{year}{2022}\natexlab{}.
\newblock \showarticletitle{Understanding Shared Autonomous Vehicle Preferences: A Comparison between Shuttles, Buses, Ridesharing and Taxis}.
\newblock \bibinfo{journal}{\emph{Sustainability}} \bibinfo{volume}{14}, \bibinfo{number}{20} (\bibinfo{date}{Oct} \bibinfo{year}{2022}), \bibinfo{pages}{13656}.
\newblock
\showISSN{2071-1050}
\urldef\tempurl%
\url{https://doi.org/10.3390/su142013656}
\showDOI{\tempurl}


\bibitem[\protect\citeauthoryear{Commission}{Commission}{2022}]%
        {TLC_Data}
\bibfield{author}{\bibinfo{person}{NYC Taxi \&~Limousine Commission}.} \bibinfo{year}{2020-2022}\natexlab{}.
\newblock \bibinfo{title}{TLC Trip Record Data}.
\newblock
\newblock
\urldef\tempurl%
\url{https://www.nyc.gov/site/tlc/about/tlc-trip-record-data.page}
\showURL{%
\tempurl}


\bibitem[\protect\citeauthoryear{COMNET}{COMNET}{2016}]%
        {Comnet2016}
\bibfield{author}{\bibinfo{person}{COMNET}.} \bibinfo{year}{2016}\natexlab{}.
\newblock \bibinfo{title}{Factsheet COMNET Overview - Appendix C: Schedules}.
\newblock
\newblock
\urldef\tempurl%
\url{https://www.comnet.org/appendix-c-schedules}
\showURL{%
\tempurl}


\bibitem[\protect\citeauthoryear{Conneau, Kiela, Schwenk, Barrault, and Bordes}{Conneau et~al\mbox{.}}{2018}]%
        {conneau2018supervised}
\bibfield{author}{\bibinfo{person}{Alexis Conneau}, \bibinfo{person}{Douwe Kiela}, \bibinfo{person}{Holger Schwenk}, \bibinfo{person}{Loic Barrault}, {and} \bibinfo{person}{Antoine Bordes}.} \bibinfo{year}{2018}\natexlab{}.
\newblock \bibinfo{title}{Supervised Learning of Universal Sentence Representations from Natural Language Inference Data}.
\newblock
\newblock
\showeprint[arxiv]{1705.02364}~[cs.CL]


\bibitem[\protect\citeauthoryear{Croes}{Croes}{1958}]%
        {Croes1958}
\bibfield{author}{\bibinfo{person}{G.~A. Croes}.} \bibinfo{year}{1958}\natexlab{}.
\newblock \showarticletitle{A Method for Solving Traveling-Salesman Problems}.
\newblock \bibinfo{journal}{\emph{Operations Research}} \bibinfo{volume}{6}, \bibinfo{number}{6} (\bibinfo{year}{1958}), \bibinfo{pages}{791--812}.
\newblock
\showISSN{0030364X, 15265463}
\urldef\tempurl%
\url{http://www.jstor.org/stable/167074}
\showURL{%
\tempurl}


\bibitem[\protect\citeauthoryear{Damle, Minden, and Ying}{Damle et~al\mbox{.}}{2018}]%
        {Damle2018}
\bibfield{author}{\bibinfo{person}{Anil Damle}, \bibinfo{person}{Victor Minden}, {and} \bibinfo{person}{Lexing Ying}.} \bibinfo{year}{2018}\natexlab{}.
\newblock \showarticletitle{{Simple, direct and efficient multi-way spectral clustering}}.
\newblock \bibinfo{journal}{\emph{Information and Inference: A Journal of the IMA}} \bibinfo{volume}{8}, \bibinfo{number}{1} (\bibinfo{date}{06} \bibinfo{year}{2018}), \bibinfo{pages}{181--203}.
\newblock
\showISSN{2049-8772}
\urldef\tempurl%
\url{https://doi.org/10.1093/imaiai/iay008}
\showDOI{\tempurl}
\showeprint{https://academic.oup.com/imaiai/article-pdf/8/1/181/28053156/iay008.pdf}


\bibitem[\protect\citeauthoryear{Delarue, Anderson, and Tjandraatmadja}{Delarue et~al\mbox{.}}{2020}]%
        {Delarue2020}
\bibfield{author}{\bibinfo{person}{Arthur Delarue}, \bibinfo{person}{Ross Anderson}, {and} \bibinfo{person}{Christian Tjandraatmadja}.} \bibinfo{year}{2020}\natexlab{}.
\newblock \showarticletitle{Reinforcement Learning with Combinatorial Actions: An Application to Vehicle Routing}. In \bibinfo{booktitle}{\emph{Advances in Neural Information Processing Systems}}, \bibfield{editor}{\bibinfo{person}{H.~Larochelle}, \bibinfo{person}{M.~Ranzato}, \bibinfo{person}{R.~Hadsell}, \bibinfo{person}{M.F. Balcan}, {and} \bibinfo{person}{H.~Lin}} (Eds.), Vol.~\bibinfo{volume}{33}. \bibinfo{publisher}{Curran Associates, Inc.}, \bibinfo{pages}{609--620}.
\newblock


\bibitem[\protect\citeauthoryear{Duan and Pettie}{Duan and Pettie}{2014}]%
        {Duan2014}
\bibfield{author}{\bibinfo{person}{Ran Duan} {and} \bibinfo{person}{Seth Pettie}.} \bibinfo{year}{2014}\natexlab{}.
\newblock \showarticletitle{Linear-Time Approximation for Maximum Weight Matching}.
\newblock \bibinfo{journal}{\emph{J. ACM}} \bibinfo{volume}{61}, \bibinfo{number}{1}, Article \bibinfo{articleno}{1} (\bibinfo{year}{2014}), \bibinfo{numpages}{23}~pages.
\newblock
\showISSN{0004-5411}
\urldef\tempurl%
\url{https://doi.org/10.1145/2529989}
\showDOI{\tempurl}


\bibitem[\protect\citeauthoryear{el~Karfi, El~Fkihi, and Faizi}{el~Karfi et~al\mbox{.}}{2018}]%
        {Karfi2018}
\bibfield{author}{\bibinfo{person}{Ikram el Karfi}, \bibinfo{person}{Sanaa El~Fkihi}, {and} \bibinfo{person}{Rdouan Faizi}.} \bibinfo{year}{2018}\natexlab{}.
\newblock \showarticletitle{A SPECTRAL CLUSTERING-BASED APPROACH FOR SENTIMENT CLASSIFICATION IN MODERN STANDARD ARABIC}. In \bibinfo{booktitle}{\emph{Proceedings of the International Conferences Big Data Analytics, Data Mining and Computational Intelligence 2018}}.
\newblock


\bibitem[\protect\citeauthoryear{Enders, Harrison, Pavone, and Schiffer}{Enders et~al\mbox{.}}{2023}]%
        {enders2023hybrid}
\bibfield{author}{\bibinfo{person}{Tobias Enders}, \bibinfo{person}{James Harrison}, \bibinfo{person}{Marco Pavone}, {and} \bibinfo{person}{Maximilian Schiffer}.} \bibinfo{year}{2023}\natexlab{}.
\newblock \showarticletitle{Hybrid multi-agent deep reinforcement learning for autonomous mobility on demand systems}. In \bibinfo{booktitle}{\emph{Learning for Dynamics and Control Conference}}. PMLR, \bibinfo{pages}{1284--1296}.
\newblock


\bibitem[\protect\citeauthoryear{Erkan and Radev}{Erkan and Radev}{2004}]%
        {erkan2004university}
\bibfield{author}{\bibinfo{person}{G{\"u}nes Erkan} {and} \bibinfo{person}{Dragomir~R Radev}.} \bibinfo{year}{2004}\natexlab{}.
\newblock \showarticletitle{The university of michigan at duc 2004}. In \bibinfo{booktitle}{\emph{Proceedings of the Document Understanding Conferences Boston, MA}}. Citeseer.
\newblock


\bibitem[\protect\citeauthoryear{Farhan and Chen}{Farhan and Chen}{2018}]%
        {Farhan2018}
\bibfield{author}{\bibinfo{person}{J. Farhan} {and} \bibinfo{person}{T.~Donna Chen}.} \bibinfo{year}{2018}\natexlab{}.
\newblock \showarticletitle{Impact of ridesharing on operational efficiency of shared autonomous electric vehicle fleet}.
\newblock \bibinfo{journal}{\emph{Transportation Research Part C: Emerging Technologies}}  \bibinfo{volume}{93} (\bibinfo{date}{08} \bibinfo{year}{2018}), \bibinfo{pages}{310--321}.
\newblock
\urldef\tempurl%
\url{https://doi.org/10.1016/j.trc.2018.04.022}
\showDOI{\tempurl}


\bibitem[\protect\citeauthoryear{Gammelli, Yang, Harrison, Rodrigues, Pereira, and Pavone}{Gammelli et~al\mbox{.}}{2022}]%
        {gammelli2022graph}
\bibfield{author}{\bibinfo{person}{Daniele Gammelli}, \bibinfo{person}{Kaidi Yang}, \bibinfo{person}{James Harrison}, \bibinfo{person}{Filipe Rodrigues}, \bibinfo{person}{Francisco Pereira}, {and} \bibinfo{person}{Marco Pavone}.} \bibinfo{year}{2022}\natexlab{}.
\newblock \showarticletitle{Graph meta-reinforcement learning for transferable autonomous mobility-on-demand}. In \bibinfo{booktitle}{\emph{Proceedings of the 28th ACM SIGKDD Conference on Knowledge Discovery and Data Mining}}. \bibinfo{pages}{2913--2923}.
\newblock


\bibitem[\protect\citeauthoryear{Garces, Bhattacharya, Gil, and Bertsekas}{Garces et~al\mbox{.}}{2023}]%
        {garces2023}
\bibfield{author}{\bibinfo{person}{Daniel Garces}, \bibinfo{person}{Sushmita Bhattacharya}, \bibinfo{person}{Stephanie Gil}, {and} \bibinfo{person}{Dimitri Bertsekas}.} \bibinfo{year}{2023}\natexlab{}.
\newblock \showarticletitle{Multiagent Reinforcement Learning for Autonomous Routing and Pickup Problem with Adaptation to Variable Demand}.
\newblock \bibinfo{journal}{\emph{International Conference on Robotics and Automation}} (\bibinfo{year}{2023}).
\newblock


\bibitem[\protect\citeauthoryear{Happle, Fonseca, and Schlueter}{Happle et~al\mbox{.}}{2019}]%
        {Happle2019}
\bibfield{author}{\bibinfo{person}{Gabriel Happle}, \bibinfo{person}{Jimeno Fonseca}, {and} \bibinfo{person}{Arno Schlueter}.} \bibinfo{year}{2019}\natexlab{}.
\newblock \bibinfo{title}{Data-driven occupancy schedules for commercial buildings}.
\newblock
\newblock
\urldef\tempurl%
\url{https://doi.org/10.3929/ethz-b-000341619}
\showDOI{\tempurl}


\bibitem[\protect\citeauthoryear{Hersbach, Bell, Biavati, Horányi, Muñoz~Sabater, Nicolas, Peubey, Radu, Rozum, Simmons, Soci, Dee, and Thépaut}{Hersbach et~al\mbox{.}}{2018}]%
        {ERA5}
\bibfield{author}{\bibinfo{person}{H. Hersbach}, \bibinfo{person}{P. Bell, B.and~Berrisford}, \bibinfo{person}{G. Biavati}, \bibinfo{person}{A. Horányi}, \bibinfo{person}{J. Muñoz~Sabater}, \bibinfo{person}{J. Nicolas}, \bibinfo{person}{C. Peubey}, \bibinfo{person}{R. Radu}, \bibinfo{person}{D. Rozum, I.and~Schepers}, \bibinfo{person}{A. Simmons}, \bibinfo{person}{C. Soci}, \bibinfo{person}{D. Dee}, {and} \bibinfo{person}{J-N. Thépaut}.} \bibinfo{year}{2018}\natexlab{}.
\newblock \bibinfo{title}{ERA5 hourly data on single levels from 1959 to present}.
\newblock
\newblock
\urldef\tempurl%
\url{https://doi.org/10.24381/cds.adbb2d47}
\showDOI{\tempurl}


\bibitem[\protect\citeauthoryear{Karp, Vazirani, and Vazirani}{Karp et~al\mbox{.}}{1990}]%
        {Karp1990}
\bibfield{author}{\bibinfo{person}{R.~M. Karp}, \bibinfo{person}{U.~V. Vazirani}, {and} \bibinfo{person}{V.~V. Vazirani}.} \bibinfo{year}{1990}\natexlab{}.
\newblock \showarticletitle{An Optimal Algorithm for On-Line Bipartite Matching}. In \bibinfo{booktitle}{\emph{Proceedings of the Twenty-Second Annual ACM Symposium on Theory of Computing}} (Baltimore, Maryland, USA) \emph{(\bibinfo{series}{STOC '90})}. \bibinfo{publisher}{Association for Computing Machinery}, \bibinfo{address}{New York, NY, USA}, \bibinfo{pages}{352–358}.
\newblock
\showISBNx{0897913612}
\urldef\tempurl%
\url{https://doi.org/10.1145/100216.100262}
\showDOI{\tempurl}


\bibitem[\protect\citeauthoryear{Li, Pan, Wu, Qi, Li, Zhang, Zhang, and Wang}{Li et~al\mbox{.}}{2012}]%
        {Li2012}
\bibfield{author}{\bibinfo{person}{Xiaolong Li}, \bibinfo{person}{Gang Pan}, \bibinfo{person}{Z. Wu}, \bibinfo{person}{Guande Qi}, \bibinfo{person}{Shijian Li}, \bibinfo{person}{Daqing Zhang}, \bibinfo{person}{Wangsheng Zhang}, {and} \bibinfo{person}{Zonghui Wang}.} \bibinfo{year}{2012}\natexlab{}.
\newblock \showarticletitle{Prediction of urban human mobility using large-scale taxi traces and its applications}.
\newblock \bibinfo{journal}{\emph{Frontiers of Computer Science in China}}  \bibinfo{volume}{6} (\bibinfo{date}{02} \bibinfo{year}{2012}), \bibinfo{pages}{111--121}.
\newblock
\urldef\tempurl%
\url{https://doi.org/10.1007/s11704-011-1192-6}
\showDOI{\tempurl}


\bibitem[\protect\citeauthoryear{Liu, Qiu, Li, Wang, Ouyang, and Lin}{Liu et~al\mbox{.}}{2019b}]%
        {Liu2019}
\bibfield{author}{\bibinfo{person}{Lingbo Liu}, \bibinfo{person}{Zhilin Qiu}, \bibinfo{person}{Guanbin Li}, \bibinfo{person}{Qing Wang}, \bibinfo{person}{Wanli Ouyang}, {and} \bibinfo{person}{Liang Lin}.} \bibinfo{year}{2019}\natexlab{b}.
\newblock \showarticletitle{Contextualized Spatial–Temporal Network for Taxi Origin-Destination Demand Prediction}.
\newblock \bibinfo{journal}{\emph{IEEE Transactions on Intelligent Transportation Systems}} \bibinfo{volume}{20}, \bibinfo{number}{10} (\bibinfo{year}{2019}), \bibinfo{pages}{3875--3887}.
\newblock
\urldef\tempurl%
\url{https://doi.org/10.1109/TITS.2019.2915525}
\showDOI{\tempurl}


\bibitem[\protect\citeauthoryear{Liu, Ott, Goyal, Du, Joshi, Chen, Levy, Lewis, Zettlemoyer, and Stoyanov}{Liu et~al\mbox{.}}{2019a}]%
        {Liu2019RoBERTaAR}
\bibfield{author}{\bibinfo{person}{Yinhan Liu}, \bibinfo{person}{Myle Ott}, \bibinfo{person}{Naman Goyal}, \bibinfo{person}{Jingfei Du}, \bibinfo{person}{Mandar Joshi}, \bibinfo{person}{Danqi Chen}, \bibinfo{person}{Omer Levy}, \bibinfo{person}{Mike Lewis}, \bibinfo{person}{Luke Zettlemoyer}, {and} \bibinfo{person}{Veselin Stoyanov}.} \bibinfo{year}{2019}\natexlab{a}.
\newblock \showarticletitle{RoBERTa: A Robustly Optimized BERT Pretraining Approach}.
\newblock \bibinfo{journal}{\emph{ArXiv}}  \bibinfo{volume}{abs/1907.11692} (\bibinfo{year}{2019}).
\newblock


\bibitem[\protect\citeauthoryear{Lowalekar, Varakantham, and Jaillet}{Lowalekar et~al\mbox{.}}{2018}]%
        {LOWALEKAR201871}
\bibfield{author}{\bibinfo{person}{Meghna Lowalekar}, \bibinfo{person}{Pradeep Varakantham}, {and} \bibinfo{person}{Patrick Jaillet}.} \bibinfo{year}{2018}\natexlab{}.
\newblock \showarticletitle{Online spatio-temporal matching in stochastic and dynamic domains}.
\newblock \bibinfo{journal}{\emph{Artificial Intelligence}}  \bibinfo{volume}{261} (\bibinfo{year}{2018}), \bibinfo{pages}{71--112}.
\newblock
\showISSN{0004-3702}
\urldef\tempurl%
\url{https://doi.org/10.1016/j.artint.2018.04.005}
\showDOI{\tempurl}


\bibitem[\protect\citeauthoryear{Markou, Kaiser, and Pereira}{Markou et~al\mbox{.}}{2019}]%
        {Markou2019}
\bibfield{author}{\bibinfo{person}{Ioulia Markou}, \bibinfo{person}{Kevin Kaiser}, {and} \bibinfo{person}{Francisco~C. Pereira}.} \bibinfo{year}{2019}\natexlab{}.
\newblock \showarticletitle{Predicting taxi demand hotspots using automated Internet Search Queries}.
\newblock \bibinfo{journal}{\emph{Transportation Research Part C: Emerging Technologies}}  \bibinfo{volume}{102} (\bibinfo{year}{2019}), \bibinfo{pages}{73--86}.
\newblock
\showISSN{0968-090X}
\urldef\tempurl%
\url{https://doi.org/10.1016/j.trc.2019.03.001}
\showDOI{\tempurl}


\bibitem[\protect\citeauthoryear{Markou, Rodrigues, and Pereira}{Markou et~al\mbox{.}}{2020}]%
        {Markou2020}
\bibfield{author}{\bibinfo{person}{Ioulia Markou}, \bibinfo{person}{Filipe Rodrigues}, {and} \bibinfo{person}{Francisco~C. Pereira}.} \bibinfo{year}{2020}\natexlab{}.
\newblock \showarticletitle{Is Travel Demand Actually Deep? An Application in Event Areas Using Semantic Information}.
\newblock \bibinfo{journal}{\emph{IEEE Transactions on Intelligent Transportation Systems}} \bibinfo{volume}{21}, \bibinfo{number}{2} (\bibinfo{year}{2020}), \bibinfo{pages}{641--652}.
\newblock
\urldef\tempurl%
\url{https://doi.org/10.1109/TITS.2019.2897341}
\showDOI{\tempurl}


\bibitem[\protect\citeauthoryear{Miao, Han, Lin, Wang, Stankovic, Hendawi, Zhang, He, and Pappas}{Miao et~al\mbox{.}}{2017}]%
        {miao2017datadriven}
\bibfield{author}{\bibinfo{person}{Fei Miao}, \bibinfo{person}{Shuo Han}, \bibinfo{person}{Shan Lin}, \bibinfo{person}{Qian Wang}, \bibinfo{person}{John Stankovic}, \bibinfo{person}{Abdeltawab Hendawi}, \bibinfo{person}{Desheng Zhang}, \bibinfo{person}{Tian He}, {and} \bibinfo{person}{George~J. Pappas}.} \bibinfo{year}{2017}\natexlab{}.
\newblock \bibinfo{title}{Data-Driven Robust Taxi Dispatch under Demand Uncertainties}.
\newblock
\newblock
\showeprint[arxiv]{1603.06263}~[cs.SY]


\bibitem[\protect\citeauthoryear{Mitchell}{Mitchell}{1998}]%
        {Mitchell1998}
\bibfield{author}{\bibinfo{person}{Melanie Mitchell}.} \bibinfo{year}{1998}\natexlab{}.
\newblock \bibinfo{booktitle}{\emph{An Introduction to Genetic Algorithms}}.
\newblock \bibinfo{publisher}{MIT Press}, \bibinfo{address}{Cambridge, MA, USA}.
\newblock
\showISBNx{0262631857}


\bibitem[\protect\citeauthoryear{Moreira-Matias, Gama, Ferreira, Mendes-Moreira, and Damas}{Moreira-Matias et~al\mbox{.}}{2013}]%
        {Moreira2013}
\bibfield{author}{\bibinfo{person}{Luis Moreira-Matias}, \bibinfo{person}{João Gama}, \bibinfo{person}{Michel Ferreira}, \bibinfo{person}{João Mendes-Moreira}, {and} \bibinfo{person}{Luis Damas}.} \bibinfo{year}{2013}\natexlab{}.
\newblock \showarticletitle{Predicting Taxi–Passenger Demand Using Streaming Data}.
\newblock \bibinfo{journal}{\emph{IEEE Transactions on Intelligent Transportation Systems}} \bibinfo{volume}{14}, \bibinfo{number}{3} (\bibinfo{year}{2013}), \bibinfo{pages}{1393--1402}.
\newblock
\urldef\tempurl%
\url{https://doi.org/10.1109/TITS.2013.2262376}
\showDOI{\tempurl}


\bibitem[\protect\citeauthoryear{Muñoz~Sabater}{Muñoz~Sabater}{2019}]%
        {ERA5-land}
\bibfield{author}{\bibinfo{person}{J. Muñoz~Sabater}.} \bibinfo{year}{2019}\natexlab{}.
\newblock \bibinfo{title}{ERA5-Land hourly data from 1981 to present}.
\newblock
\newblock
\urldef\tempurl%
\url{https://doi.org/10.24381/cds.e2161bac}
\showDOI{\tempurl}


\bibitem[\protect\citeauthoryear{Nazari, Oroojlooy, Snyder, and Takac}{Nazari et~al\mbox{.}}{2018}]%
        {Nazari2018}
\bibfield{author}{\bibinfo{person}{MohammadReza Nazari}, \bibinfo{person}{Afshin Oroojlooy}, \bibinfo{person}{Lawrence Snyder}, {and} \bibinfo{person}{Martin Takac}.} \bibinfo{year}{2018}\natexlab{}.
\newblock \showarticletitle{Reinforcement Learning for Solving the Vehicle Routing Problem}. In \bibinfo{booktitle}{\emph{Advances in Neural Information Processing Systems}}, \bibfield{editor}{\bibinfo{person}{S.~Bengio}, \bibinfo{person}{H.~Wallach}, \bibinfo{person}{H.~Larochelle}, \bibinfo{person}{K.~Grauman}, \bibinfo{person}{N.~Cesa-Bianchi}, {and} \bibinfo{person}{R.~Garnett}} (Eds.), Vol.~\bibinfo{volume}{31}. \bibinfo{publisher}{Curran Associates, Inc.}
\newblock
\urldef\tempurl%
\url{https://proceedings.neurips.cc/paper_files/paper/2018/file/9fb4651c05b2ed70fba5afe0b039a550-Paper.pdf}
\showURL{%
\tempurl}


\bibitem[\protect\citeauthoryear{Nejadettehad, Mahini, and Bahrak}{Nejadettehad et~al\mbox{.}}{2019}]%
        {Nejadettehad2019}
\bibfield{author}{\bibinfo{person}{Alireza Nejadettehad}, \bibinfo{person}{Hamid Mahini}, {and} \bibinfo{person}{Behnam Bahrak}.} \bibinfo{year}{2019}\natexlab{}.
\newblock \showarticletitle{Short-term Demand Forecasting for Online Car-hailing Services Using Recurrent Neural Networks}.
\newblock \bibinfo{journal}{\emph{Applied Artificial Intelligence}}  \bibinfo{volume}{34} (\bibinfo{year}{2019}), \bibinfo{pages}{674 -- 689}.
\newblock


\bibitem[\protect\citeauthoryear{Open-Meteo}{Open-Meteo}{2022}]%
        {Open-Meteo}
\bibfield{author}{\bibinfo{person}{Open-Meteo}.} \bibinfo{year}{2022}\natexlab{}.
\newblock \bibinfo{title}{Historical Weather API}.
\newblock
\newblock
\urldef\tempurl%
\url{https://open-meteo.com/en/docs/historical-weather-api}
\showURL{%
\tempurl}


\bibitem[\protect\citeauthoryear{Panchabikesan, Haghighat, and Mankibi}{Panchabikesan et~al\mbox{.}}{2021}]%
        {PANCHABIKESAN2021119539}
\bibfield{author}{\bibinfo{person}{Karthik Panchabikesan}, \bibinfo{person}{Fariborz Haghighat}, {and} \bibinfo{person}{Mohamed~El Mankibi}.} \bibinfo{year}{2021}\natexlab{}.
\newblock \showarticletitle{Data driven occupancy information for energy simulation and energy use assessment in residential buildings}.
\newblock \bibinfo{journal}{\emph{Energy}}  \bibinfo{volume}{218} (\bibinfo{year}{2021}), \bibinfo{pages}{119539}.
\newblock
\showISSN{0360-5442}
\urldef\tempurl%
\url{https://doi.org/10.1016/j.energy.2020.119539}
\showDOI{\tempurl}


\bibitem[\protect\citeauthoryear{{Parvez Farazi}, Zou, Ahamed, and Barua}{{Parvez Farazi} et~al\mbox{.}}{2021}]%
        {PARVEZFARAZI2021100425}
\bibfield{author}{\bibinfo{person}{Nahid {Parvez Farazi}}, \bibinfo{person}{Bo Zou}, \bibinfo{person}{Tanvir Ahamed}, {and} \bibinfo{person}{Limon Barua}.} \bibinfo{year}{2021}\natexlab{}.
\newblock \showarticletitle{Deep reinforcement learning in transportation research: A review}.
\newblock \bibinfo{journal}{\emph{Transportation Research Interdisciplinary Perspectives}}  \bibinfo{volume}{11} (\bibinfo{year}{2021}), \bibinfo{pages}{100425}.
\newblock
\showISSN{2590-1982}
\urldef\tempurl%
\url{https://doi.org/10.1016/j.trip.2021.100425}
\showDOI{\tempurl}


\bibitem[\protect\citeauthoryear{Platform}{Platform}{2022}]%
        {Google_Maps}
\bibfield{author}{\bibinfo{person}{Google~Maps Platform}.} \bibinfo{year}{2022}\natexlab{}.
\newblock \bibinfo{title}{Nearby Search API}.
\newblock
\newblock
\urldef\tempurl%
\url{https://developers.google.com/maps/documentation/places/web-service/search-nearby}
\showURL{%
\tempurl}


\bibitem[\protect\citeauthoryear{PredictHQ}{PredictHQ}{2022}]%
        {PHQ}
\bibfield{author}{\bibinfo{person}{PredictHQ}.} \bibinfo{year}{2022}\natexlab{}.
\newblock \bibinfo{title}{Attended events API}.
\newblock
\newblock
\urldef\tempurl%
\url{https://www.predicthq.com/apis/event-api}
\showURL{%
\tempurl}


\bibitem[\protect\citeauthoryear{Rodrigues, Markou, and Pereira}{Rodrigues et~al\mbox{.}}{2019}]%
        {Rodrigues2019}
\bibfield{author}{\bibinfo{person}{Filipe Rodrigues}, \bibinfo{person}{Ioulia Markou}, {and} \bibinfo{person}{Francisco~C. Pereira}.} \bibinfo{year}{2019}\natexlab{}.
\newblock \showarticletitle{Combining time-series and textual data for taxi demand prediction in event areas: A deep learning approach}.
\newblock \bibinfo{journal}{\emph{Information Fusion}}  \bibinfo{volume}{49} (\bibinfo{year}{2019}), \bibinfo{pages}{120--129}.
\newblock
\showISSN{1566-2535}
\urldef\tempurl%
\url{https://doi.org/10.1016/j.inffus.2018.07.007}
\showDOI{\tempurl}


\bibitem[\protect\citeauthoryear{SerAPI}{SerAPI}{2022}]%
        {SerpAPI}
\bibfield{author}{\bibinfo{person}{SerAPI}.} \bibinfo{year}{2022}\natexlab{}.
\newblock \bibinfo{title}{Google Maps API}.
\newblock
\newblock
\urldef\tempurl%
\url{https://serpapi.com/google-maps-api}
\showURL{%
\tempurl}


\bibitem[\protect\citeauthoryear{Shao, Kumar, and Fletcher}{Shao et~al\mbox{.}}{2017}]%
        {shao2017riemannian}
\bibfield{author}{\bibinfo{person}{Hang Shao}, \bibinfo{person}{Abhishek Kumar}, {and} \bibinfo{person}{P.~Thomas Fletcher}.} \bibinfo{year}{2017}\natexlab{}.
\newblock \bibinfo{title}{The Riemannian Geometry of Deep Generative Models}.
\newblock
\newblock
\showeprint[arxiv]{1711.08014}~[cs.LG]


\bibitem[\protect\citeauthoryear{Silver and Veness}{Silver and Veness}{2010}]%
        {SiV10}
\bibfield{author}{\bibinfo{person}{David Silver} {and} \bibinfo{person}{Joel Veness}.} \bibinfo{year}{2010}\natexlab{}.
\newblock \showarticletitle{Monte-{Carlo Planning in Large POMDP}s}. In \bibinfo{booktitle}{\emph{Proc. 23rd International Conf. on NeurIPS}} (Vancouver, British Columbia, Canada). \bibinfo{address}{Red Hook, NY, USA}, \bibinfo{pages}{2164--2172}.
\newblock


\bibitem[\protect\citeauthoryear{Subramanian, Trischler, Bengio, and Pal}{Subramanian et~al\mbox{.}}{2018}]%
        {subramanian2018learning}
\bibfield{author}{\bibinfo{person}{Sandeep Subramanian}, \bibinfo{person}{Adam Trischler}, \bibinfo{person}{Yoshua Bengio}, {and} \bibinfo{person}{Christopher~J Pal}.} \bibinfo{year}{2018}\natexlab{}.
\newblock \bibinfo{title}{Learning General Purpose Distributed Sentence Representations via Large Scale Multi-task Learning}.
\newblock
\newblock
\showeprint[arxiv]{1804.00079}~[cs.CL]


\bibitem[\protect\citeauthoryear{Ulmer, Goodson, Mattfeld, and Hennig}{Ulmer et~al\mbox{.}}{2019}]%
        {Ulmer2018}
\bibfield{author}{\bibinfo{person}{Marlin~W. Ulmer}, \bibinfo{person}{Justin~C. Goodson}, \bibinfo{person}{Dirk~C. Mattfeld}, {and} \bibinfo{person}{Marco Hennig}.} \bibinfo{year}{2019}\natexlab{}.
\newblock \showarticletitle{Offline–Online Approximate Dynamic Programming for Dynamic Vehicle Routing with Stochastic Requests}.
\newblock \bibinfo{journal}{\emph{Transportation Science}} \bibinfo{volume}{53}, \bibinfo{number}{1} (\bibinfo{year}{2019}), \bibinfo{pages}{185--202}.
\newblock
\urldef\tempurl%
\url{https://doi.org/10.1287/trsc.2017.0767}
\showDOI{\tempurl}
\showeprint{https://doi.org/10.1287/trsc.2017.0767}


\bibitem[\protect\citeauthoryear{Werger and Matari{\'c}}{Werger and Matari{\'c}}{2000}]%
        {werger2000broadcast}
\bibfield{author}{\bibinfo{person}{Barry~Brian Werger} {and} \bibinfo{person}{Maja~J Matari{\'c}}.} \bibinfo{year}{2000}\natexlab{}.
\newblock \showarticletitle{Broadcast of local eligibility for multi-target observation}.
\newblock In \bibinfo{booktitle}{\emph{Distributed Autonomous Robotic Systems 4}}. \bibinfo{publisher}{Springer}, \bibinfo{pages}{347--356}.
\newblock


\bibitem[\protect\citeauthoryear{Yang, Yang, Cer, Law, and Darve}{Yang et~al\mbox{.}}{2021}]%
        {yang-etal-2021-universal}
\bibfield{author}{\bibinfo{person}{Ziyi Yang}, \bibinfo{person}{Yinfei Yang}, \bibinfo{person}{Daniel Cer}, \bibinfo{person}{Jax Law}, {and} \bibinfo{person}{Eric Darve}.} \bibinfo{year}{2021}\natexlab{}.
\newblock \showarticletitle{Universal Sentence Representation Learning with Conditional Masked Language Model}. In \bibinfo{booktitle}{\emph{Proceedings of the 2021 Conference on Empirical Methods in Natural Language Processing}}. \bibinfo{publisher}{Association for Computational Linguistics}, \bibinfo{address}{Online and Punta Cana, Dominican Republic}, \bibinfo{pages}{6216--6228}.
\newblock
\urldef\tempurl%
\url{https://doi.org/10.18653/v1/2021.emnlp-main.502}
\showDOI{\tempurl}


\bibitem[\protect\citeauthoryear{Yannakakis, Tovey, Korst, and J.~M.~van Laarhoven}{Yannakakis et~al\mbox{.}}{2003}]%
        {LSCP2003}
\bibfield{author}{\bibinfo{person}{Mihalis Yannakakis}, \bibinfo{person}{Craig~A. Tovey}, \bibinfo{person}{Jan H.~M. Korst}, {and} \bibinfo{person}{Jan H.~M. J.~M.~van Laarhoven}.} \bibinfo{year}{2003}\natexlab{}.
\newblock \bibinfo{booktitle}{\emph{Local Search in Combinatorial Optimization}}.
\newblock \bibinfo{publisher}{Princeton University Press}.
\newblock
\urldef\tempurl%
\url{http://www.jstor.org/stable/j.ctv346t9c}
\showURL{%
\tempurl}


\bibitem[\protect\citeauthoryear{Zhu, Galstyan, Cheng, and Lerman}{Zhu et~al\mbox{.}}{2014}]%
        {zhu2014tripartite}
\bibfield{author}{\bibinfo{person}{Linhong Zhu}, \bibinfo{person}{Aram Galstyan}, \bibinfo{person}{James Cheng}, {and} \bibinfo{person}{Kristina Lerman}.} \bibinfo{year}{2014}\natexlab{}.
\newblock \showarticletitle{Tripartite graph clustering for dynamic sentiment analysis on social media}. In \bibinfo{booktitle}{\emph{Proceedings of the 2014 ACM SIGMOD international conference on Management of data}}. \bibinfo{pages}{1531--1542}.
\newblock


\bibitem[\protect\citeauthoryear{Zhu, Gao, Pan, Li, Deng, and Shahabi}{Zhu et~al\mbox{.}}{2013}]%
        {zhu2013graph}
\bibfield{author}{\bibinfo{person}{Linhong Zhu}, \bibinfo{person}{Sheng Gao}, \bibinfo{person}{Sinno~Jialin Pan}, \bibinfo{person}{Haizhou Li}, \bibinfo{person}{Dingxiong Deng}, {and} \bibinfo{person}{Cyrus Shahabi}.} \bibinfo{year}{2013}\natexlab{}.
\newblock \showarticletitle{Graph-based informative-sentence selection for opinion summarization}. In \bibinfo{booktitle}{\emph{Proceedings of the 2013 IEEE/ACM International Conference on Advances in Social Networks Analysis and Mining}}. \bibinfo{pages}{408--412}.
\newblock


\bibitem[\protect\citeauthoryear{Zimbra, Abbasi, Zeng, and Chen}{Zimbra et~al\mbox{.}}{2018}]%
        {Zimbra2018}
\bibfield{author}{\bibinfo{person}{David Zimbra}, \bibinfo{person}{Ahmed Abbasi}, \bibinfo{person}{Daniel Zeng}, {and} \bibinfo{person}{Hsinchun Chen}.} \bibinfo{year}{2018}\natexlab{}.
\newblock \showarticletitle{The State-of-the-Art in Twitter Sentiment Analysis: A Review and Benchmark Evaluation}.
\newblock \bibinfo{journal}{\emph{ACM Trans. Manage. Inf. Syst.}} \bibinfo{volume}{9}, \bibinfo{number}{2}, Article \bibinfo{articleno}{5} (\bibinfo{date}{aug} \bibinfo{year}{2018}), \bibinfo{numpages}{29}~pages.
\newblock
\showISSN{2158-656X}
\urldef\tempurl%
\url{https://doi.org/10.1145/3185045}
\showDOI{\tempurl}


\end{thebibliography}


\end{document}